\documentclass[11pt,DIV=12,BCOR=0.7cm]{article}   

\usepackage[left=2cm, right=2cm, top=2cm, bottom=3cm]{geometry}
\usepackage{graphicx,amssymb}
\usepackage{amsmath,amsfonts, amsthm}
\usepackage{bm}
\usepackage{mathtools,mathdots}
\usepackage{pgfplots}
    \pgfplotsset{
	layers/my layer set/.define layer set={
		background,
		main,
		foreground
	}{
	},
	set layers=my layer set,
}
\usepackage{subcaption}
\usepackage{algorithm,algorithmic}
\usepackage{setspace}
\usepackage{xcolor}
\usepackage{scalerel,stackengine}
\usepackage{mathtools}
\usepackage{tikz}
\usetikzlibrary{matrix,chains,positioning,decorations.pathreplacing,arrows}
\usetikzlibrary{positioning,calc}
\usepackage{url}
\usepackage{wrapfig, multirow}
\usepackage{tablefootnote}
\usepackage{textcomp}
\usepackage[hidelinks]{hyperref}
\usepackage{makecell}
\usepackage{arydshln}
\usepackage{accents}
\usepackage{pifont}
\usepackage[normalem]{ulem}

\DeclarePairedDelimiterX{\infdivx}[2]{\big(}{\big)}{%
	#1\;\delimsize\|\;#2%
}

\DeclarePairedDelimiter{\norm}\lVert\rVert

\DeclareMathOperator{\E}{\mathbb{E}}

\newcommand{\expnumber}[2]{{#1}\mathrm{e}{#2}}

\newcommand{\cmark}{\ding{51}}%
\newcommand{\xmark}{\ding{55}}%

\newenvironment{keywords}
{\bgroup\leftskip 20pt\rightskip 20pt \small\noindent{\bf Keywords:} }%
{\par\egroup\vskip 0.25ex}

\makeatletter
\DeclareRobustCommand{\Udots}{%
	\vcenter{\offinterlineskip
		\halign{%
			\hbox to .8em{##}\cr
			\hfil.\cr\noalign{\kern.2ex}
			\hfil.\hfil\cr\noalign{\kern.2ex}
			.\hfil\cr}%
	}%
}
\makeatother

\title{Online Training and Pruning of Deep Reinforcement Learning Networks}

\author{Valentin Frank Ingmar Guenter\thanks{Email:
vgunter@uci.edu}\ \ and\ \ Athanasios~Sideris\thanks{Email:
asideris@uci.edu}\\ \\  Department of Mechanical and
Aerospace Engineering,\\ University of California, Irvine,\\
Irvine, CA, 92697}
\date{}
\begin{document}
				
\maketitle
\begin{abstract}
Scaling deep neural networks (NN) of reinforcement learning (RL) algorithms has been shown to enhance performance when feature extraction networks are used but the gained performance comes at the significant expense of increased computational and memory complexity. Neural network pruning methods have successfully addressed this challenge in supervised learning.
However, their application to RL is underexplored.
We propose an approach to integrate simultaneous training and pruning within advanced RL methods, in particular to RL algorithms enhanced by the Online Feature Extractor Network (OFENet).
Our networks (\textit{XiNet}) are trained to solve stochastic optimization problems over the RL networks' weights and the parameters of variational Bernoulli distributions for 0/1 Random Variables $\xi$ scaling each unit in the networks. The stochastic problem formulation induces regularization terms that promote convergence of the variational parameters to 0 when a unit contributes little to the performance. In this case, the corresponding structure is rendered permanently inactive and pruned from its network.
We propose a cost-aware, sparsity-promoting regularization scheme, tailored to the DenseNet architecture of OFENets expressing the parameter complexity of involved networks in terms of the parameters of the RVs in these networks. Then, when matching this cost with the regularization terms, the many hyperparameters associated with them are automatically selected, effectively combining the RL objectives and network compression. We evaluate our method on continuous control benchmarks (MuJoCo) and the Soft Actor-Critic RL agent, demonstrating that OFENets can be pruned considerably with minimal loss in performance. Furthermore, our results confirm that pruning large networks during training produces more efficient and higher performing RL agents rather than training smaller networks from scratch.
\end{abstract}


\section{Introduction}

Large-scale deep neural networks (DNN) have achieved state-of-the-art results across a wide variety of domains, including computer vision \cite{krizhevsky2012imagenet}, object classification \cite{chen2017rethinking}, speech recognition \cite{graves2013speech}, and natural language processing \cite{vaswani2017attention}. Beyond these traditional supervised learning tasks, neural networks (NN) have also become indispensable in other machine learning paradigms, such as self-supervised learning \cite{liu2021self, brown2020language} and reinforcement learning (RL) \cite{mnih_playing_2013, mnih_human-level_2015}. In particular, DNN are used in RL as a means to learn and execute one or more of the various functions such as the agent's policy, a model of the environment and the Value function, Action-Value and Advantage functions as required by the particular RL method.

In contrast to supervised learning, simply scaling up the DNNs in RL implementations \cite{SAC_haarnoja2018soft,TD3_fujimoto2018addressing,PPO_schulman_proximal_2017} often leads to training instabilities that adversely impact overall performance \cite{vanhasselt2018, sinha2020, andrychowicz2020matters, silver2014deterministic}.
Moreover, the selection of a suitable DNN architecture and extensive manual tuning of associated hyperparameters further complicate efforts to train and deploy RL agents \cite{andrychowicz2020matters}.
Despite these challenges, recent research has demonstrated that with the right techniques scaling up networks in RL applications can significantly enhance performance and allow RL agents to achieve higher scores on demanding benchmarks \cite{ota2020can, ota_framework_2024, schwarzer_bff, protovaluenet_farebrother2023, hafner2023mastering}. However, this comes at the cost of often tremendously increased computational complexity.

DNN complexity has motivated the use of DNN pruning approaches that have been shown to achieve in many cases considerable reduction in the complexity of the network while sacrificing only a little of its performance. Such neural network pruning \cite{han_learning_2015, blalock_what_2020} has proven to be an effective approach for reducing the computational and memory footprint of DNNs in supervised learning scenarios. Initially, pruning methods focused on eliminating individual weights \cite{han_learning_2015}, but more recent advances have established that structured unit and filter pruning \cite{li_pruning_2017, liebenwein_lost_2021, SCOP_tang_2020, Guenter24_Robust}, as well as layer pruning \cite{chen19_shallowing, DBPwang2019, guenter2024concurrenttraininglayerpruning}, yield more substantial efficiency gains. Furthermore, techniques that combine training and pruning from the outset can provide computational savings during both the training and deployment phases by directly fostering compact models \cite{Guenter24_Robust, guenter2024concurrenttraininglayerpruning}. Notably, these pruning techniques often find that training large networks and subsequently pruning them yields superior results compared to training smaller networks from scratch \cite{blalock_what_2020, frankle_lottery_2019, guenter2024concurrenttraininglayerpruning}.

While most pruning approaches were originally developed within the context of supervised learning, there is a growing body of work exploring their effectiveness within RL \cite{graesser2022state, Grooten23_filtering, Ceron24_invaluebased, guenter_phdthesis}. In particular, the question of whether simultaneous training and pruning methods can be successfully utilized with modern RL algorithms employing large-scale DNNs remains largely underexplored.
Motivated by this situation, we investigate in this work how the simultaneous training and pruning approach of \cite{guenter2024complexityawaretrainingdeepneural, Guenter24_Robust}, which has been shown to be effective in supervised learning settings can be adapted and integrated with contemporary RL methods, particularly those capable of scaling to and leveraging the advantages of large DNN architectures. The goal is to start training with initially  large networks and prune them down during training to high performing smaller models when compared to those used in traditional RL methods. An important characteristic of the approach of \cite{guenter2024complexityawaretrainingdeepneural, Guenter24_Robust}, which suggests its suitability for this purpose, is that during training only smaller subnetworks are explored that potentially mitigates the brittleness of DNN training in RL configurations \cite{song2019observational, zhang2018studyoverfitting}.
In particular, we consider RL architectures enhanced with the OFENet (Online Feature Extractor Network)  introduced in  \cite{ota2020can} based on the \textit{model network} architecture of \cite{munk2016learning} that can be added to many existing and established RL algorithms and allow them to effectively use larger DNNs models.  The function of OFENet is to extract a rich set of features from the observations made in the environment (measured state) and the actions effected by the RL agent and use them as inputs to the networks implementing policy and value functions. These features are obtained by training a DenseNet \cite{huang2017densely} and include the original observations and actions thus introducing an apparent redundancy. However, \cite{ota2020can} shows that this redundancy allows to scale up the DNN involved while maintaining stable training and can substantially increase the performance of RL agents.
Subsequently, \cite{ota_framework_2024} extended this work by using distributed and parallel data acquisition in RL algorithms to improve performance and capacity for scaling even further. However, due to the much larger amount of collected data the method becomes less sample-efficient and computationally even more demanding.
The increase of performance with the use of OFENet comes at the cost of excessive network complexity due to both the OFENet structure and the scaled-up RL networks in each method. Therefore, OFENets present an appealing circumstance to apply pruning techniques such as the ones in \cite{guenter2024complexityawaretrainingdeepneural, Guenter24_Robust} to considerably reduce the size of the NNs while preserving their enhanced performance.

\subsection*{Summary of Contributions}
\begin{itemize}
\item We integrate the simultaneous training and pruning approach of \cite{guenter2024complexityawaretrainingdeepneural, Guenter24_Robust} with any RL algorithm that employs DNNs and in particular OFENets to learn value functions, and/or a control policy, and/or a dynamic model of the environment and investigate their efficacy in obtaining networks of smaller sizes while maintaining a high performance score compared to traditional RL algorithms and their associated networks. We refer to the so modified OFENet as the OFEXiNet.

\item
We define a sparsity promoting and complexity-aware regularization cost that combines with the RL training goals to trade-off network performance and computational complexity and proves to be crucial for the balanced pruning of the DenseNet \cite{huang2017densely} architecture of OFENets.
\item
We apply our approach to popular robotic control environments provided by MuJoCo\footnote{https://mujoco.org/} with the \textit{Gymnasium} suite \cite{gymnasium2024towers} and the RL agent SAC \cite{SAC_haarnoja2018soft}. Our method is compared to the original SAC method without the use of OFENets and the results of \cite{ota2020can} where OFENets are used.
We demonstrate that the large NN models in RL algorithms using OFENets can be reduced down to $40\%$ of their initial size with only little performance loss. Furthermore, greater reduction is obtained for the part of the OFENet required in the implementation of the policy and for the deployment of the agent.
Similar to the supervised learning setting, we find that starting from larger networks and pruning them to suitable size yields higher performing agents of similar size when compared to traditional RL methods.
\end{itemize}

The remainder of the paper is organized as follows.
In Section~\ref{sec:background}, we provide the necessary background on simultaneous training and pruning of DNNs (Section~\ref{sec:pruning}) and OFE networks (Section~\ref{sec:OFE}).
Section~\ref{sec:RL} lays out the steps for the integration of simultaneous training and pruning of OFE networks and the DNN models for existing RL algorithms.
In Section~\ref{sec:ofe-xi}, we present the equations underlying the architecture and the optimization objective for the training the OFEXiNet and in Section~\ref{sec:comp_cmpx} we give the details behind the proposed sparsity promoting and cost-aware regularization term in the optimization objective.
Section~\ref{sec:algo} summarizes our proposed algorithm for simultaneous RL and pruning using the OFEXiNet.
Section~\ref{sec:relwork} discusses some alternative pruning approaches for RL networks.
In Section~\ref{sec:simulations}, we provide extensive simulation results for popular robotic control environments (MuJoCo with \textit{Gymnasium} \cite{gymnasium2024towers}) and RL agents and demonstrate the significant reduction in complexity of the RL networks that can be obtained with our method.
Lastly, Section~\ref{sec:conclusion} summarizes and concludes the paper.

\paragraph{Notation:}
We use the subscript $l$ to distinguish parameters or variables of the $l$th layer of a neural network.
Also $\norm{\cdot}$ denotes the Euclidean norm (2-norm) of a vector and $\norm{\cdot}_1$ denotes the 1-norm of a vector. $\E\left[\cdot\right]$ denotes taking expectation with respect to the indicated random variables. We introduce other notations before their use.

\section{Background Material}\label{sec:background}

\subsection{Simultaneous Training and Pruning of Deep Neural Networks (XiNet)}\label{sec:pruning}
Next, we briefly review the main ideas in the approach of  \cite{Guenter24_Robust, guenter2024complexityawaretrainingdeepneural} for the simultaneous training and pruning of neural networks.
In this approach, binary ($0/1$) Random Variables (RV) $\xi_k$ with assumed variational posterior Bernoulli distributions $\xi_k\sim Bernoulli(\theta_k)$, $0\leq\theta_k\leq 1$ are introduced multiplying the (scalar) outputs of each unit in the hidden layers considered for elimination. We then refer to the NN modified in this manner as a {\it XiNet}. In the following, we collectively denote the $\xi_k$ and $\theta_k$ as $\Xi$ and $\Theta$, respectively.
Based on a statistical formulation, a cost function is considered that combines minimization of the expected value of the training error over the data and the $\Xi$ RVs, standard $L_2$ regularization terms on the network weights, collectively denoted as $W$, and additional regularization terms on the $\Theta$ parameters. More specifically,
given input patterns $x_i \in \mathbb{R}^m$ and corresponding targets $y_i \in \mathbb{R}^n$ forming the dataset $\mathcal{D}=\lbrace (x_i,y_i)\rbrace_{i=1}^N$ of size $N>0$, a Maximum A Posteriori formulation leads to minimize over $W$, $\Theta$:
\begin{align}\label{eq:L_train_obj}
\begin{split}
L(W,\Theta)\equiv
C(W,\Theta) + \frac{\lambda}{2}W^\top W + \Gamma_{reg}(\Theta)
\end{split}
\end{align}
where
\begin{align}\label{eq:Cprn}
\Gamma_{reg}(\Theta)\equiv -\sum_{k}\theta_k \log\gamma_k
\end{align}
\begin{align}\label{eq:Cost}
C(W,\Theta) \equiv \E_{\substack{\Xi\sim  q(\Xi\mid \Theta) \\ (x_i,y_i) \sim p(\mathcal{D})}}\left[-\log p(y_i\mid x_i,W, \Xi)\right],
\end{align}
$p(y_i\mid x_i,W,\Xi)$ represents a statistical model for the data and $p(\mathcal{D})$ is the empirical distribution, which assigns probability $\frac{1}{N}$ to each sample $(x_i, y_i)$. In \eqref{eq:L_train_obj}, $\lambda>0$ and in \eqref{eq:Cprn}, $0<\gamma_k<1$ are hyperparameters associated with the prior distributions on $W$ and $\xi_k$, respectively. In particular, key to the method is the use of the {\it flattening} hyperprior introduced in \cite{Guenter24_Robust} and defined on the parameters $\pi_k$ of prior Bernoulli distributions imposed on $\xi_k\sim Bernoulli(\pi_k)$ as
\begin{align*}
p(\pi_k \mid \gamma_k) = \frac{\gamma_k-1}{\log\gamma_k}\cdot\frac{1}{1+(\gamma_k-1)(1-\pi_k)}.
\end{align*}
It is shown in \cite{Guenter24_Robust} where unit-only pruning is considered that the flattening hyperprior leads in regularization terms $\Gamma_{reg}(\Theta)$ in \eqref{eq:L_train_obj}, which are linear in $\theta_k$ with corresponding contribution to the derivative $\frac{\partial L(W,\Theta)}{\partial \theta_k}$ equal to $-\log\gamma_k$, which is independent of $\theta_k$, i.e., flat and easy to adjust via the hyperparameter $\log\gamma_k$.
Furthermore, we note that $C(W,\Theta)$ can be expressed as
\begin{align}\label{eq:C_def}
\begin{split}
C(W,\theta_k,\bar\Theta_k) =  \theta_k C_k^1(W,\bar\Theta_k) + (1-\theta_k)C_k^0(W,\bar\Theta_k)
\end{split}
\end{align}
where
\begin{align}\label{eq:C10_def}
C_k^i\equiv \E_{\substack{\bar\Xi_k\sim  q(\bar\Xi_k\mid \bar\Theta_k) \\ (x_i,y_i) \sim p(\mathcal{D})}}\left[-\log p(y_i\mid x_i,W, \xi_k=i, \bar\Xi_k)\right],\ \ \ i=0,1
\end{align}
and $\bar\Xi_k$, $\bar\Theta_k$ denote the $\Xi$, $\Theta$ variables excluding $\xi_k$, $\theta_k$, respectively. Thus, $C^0_k$, $C^1_k$ are the expected cost when $\xi_k=0$, $\xi_k=1$, respectively and clearly $C(W,\Theta)$ is also a linear function of the $\theta_k$ parameters.

Typically, a common value $\log\gamma$ is used for all units in the network to facilitate tuning. For more flexibility, different such hyperparameters should be considered for the units in each layer in the network, which may be difficult to do in Deep Neural Networks with many layers or layers of different type and/or width. For such cases, \cite{guenter2024complexityawaretrainingdeepneural} offers an interpretation of the $\theta$ regularization term as the parameter and computational complexity of the network by which the $\log\gamma_k$ hyperparameters associated with $\theta_k$ are chosen automatically as linear functions of the remaining $\theta_j,\, j\neq k$ parameters and three additional parameters with clear meaning that makes them straightforward to tune; thus minimization of the $\theta$ regularization terms is directly tied to minimizing network complexity. A similar approach is employed here in Section~\ref{sec:comp_cmpx}.

Furthermore, we remark that in  the approach of \cite{Guenter24_Robust} and also the combined unit and layer pruning setting of \cite{guenter2024complexityawaretrainingdeepneural},  the $\theta$ regularization terms  assume a multilinear form in $\theta_k$.  Then, as it is  shown in \cite{guenter2024concurrenttraininglayerpruning, guenter2024complexityawaretrainingdeepneural}, the optimal $\theta_k^*$ are either $0$, or $1$ resulting in a deterministic network comprising the units corresponding to $\theta_k^*=1$.

During training, the cost function is minimized via stochastic gradient descent over the network weights and the $\Theta$ parameters. More specifically, a mini-batch from the available data is considered as usual and also a realization  $\hat\Xi$ is sampled and used to estimate the expectations $C(W,\Theta)$ and either $C_k^0(W,\bar\Theta_k)$ or $C_k^1(W,\bar\Theta_k)$ depending on whether $\xi_k=0$ or $\xi_k=1$, respectively; this amounts to training the sub-network (i.e., learning its weights) defined by the units for which $\hat\xi_k=1$ on the mini-batch via backpropagation. Also, $\theta_k$ is updated via projected gradient descent by
\begin{align}\label{eq:theta_update}
\begin{split}
\delta\theta_k&=-\eta \frac{\partial L(W,\Theta)}{\partial\theta_k}=-\eta\left[\frac{\partial C(W,\Theta)}{\partial\theta_k}+\frac{\partial \Gamma_{reg}(\Theta)}{\partial\theta_k}\right]
=-\eta\left[C_k^1-C_k^0+\frac{\partial \Gamma_{reg}(\Theta)}{\partial\theta_k}\right],\\[1em]
\theta_k& \,\leftarrow\, \min\left\{\max\left\{ \theta_k +\delta\theta_k,\, 0 \right\},\, 1\right\}
\end{split}
\end{align}
where $\eta>0$ is the step size and \eqref{eq:theta_update} immediately follows from \eqref{eq:L_train_obj} and \eqref{eq:C_def}.
It is interesting to note here that $C_k^1-C_k^0$ can be thought as a measure of how ``useful'' is the $k$th unit in reducing the cost.  As it was remarked above either an estimate of $C_k^0(W,\bar\Theta_k)$ or $C_k^1(W,\bar\Theta_k)$ is available. We compute the other via a first-order Taylor series approximation that amounts to
\begin{align}\label{eq:deltaC}
C_k^1-C_k^0 \approx \frac{\partial C(W,\Theta)}{\partial\xi_k},
\end{align}
which is readily available from the backpropagation. This approximation turns out to be equivalent in form with the Straight-Through estimator of \cite{bengio2013estimating}; it has been found to work well in our application and some analytical justification explaining the empirical evidence is given in \cite{Guenter24_Robust}.
Therefore, updating the $\Theta$ parameters costs little over standard backpropagation and in most cases the resulting computational savings due to network reduction during training outweighs this overhead \cite{guenter2024complexityawaretrainingdeepneural}. Indeed, smaller subnetworks are trained throughout due to the sampling process and also once some $\theta_k$ has converged to less than a specified small tolerance, its $k$th unit is eliminated and training continues with a smaller network.

\subsection{Online Feature Extractor Network (OFENet) and its Application to Reinforcement Learning} \label{sec:OFE}

\begin{figure}[t!]
	\centering
	\includegraphics[width=0.75\textwidth]{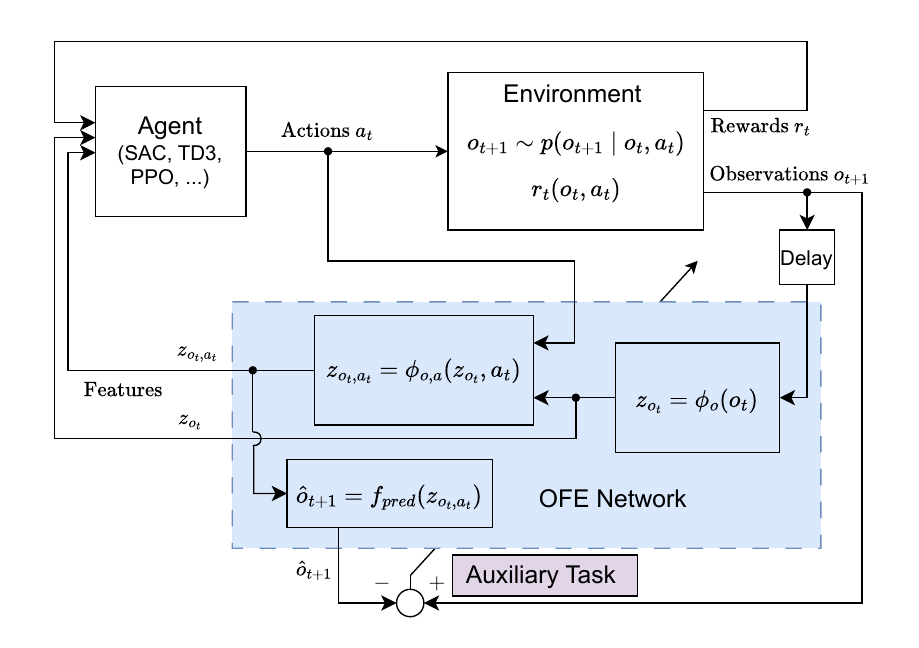}		
	\caption{Schematic integration of the Online Feature Extractor Network $\phi_{o_t}(o), \phi_{o,a}(z_{o_t},a_t)$ (OFENet, blue shaded box) in the fully observed reinforcement learning setting. The OFENet is learned online and according to the auxiliary task (purple shaded box).}\label{fig:oferl}
\end{figure}

Figure~\ref{fig:oferl} shows a discrete-time and fully observed RL-scheme with the integration of OFENet. Given the current state/observation $o_t$, the agent samples/selects an action $a_t\sim\pi(a_t\mid o_t)$ according to its control policy defined by the probability distribution $\pi$ and applies it to the environment.
Then, the environment returns the next state/observation $o_{t+1}\sim p(o_{t+1}\mid o_t, a_t)$ based on dynamics described by the probability distribution $p$ and also returns the reward $r_t$. In this manner, state-action trajectories with data pairs $o_{t_n}, a_{t_n}$, $n=1,\dots,N$ are generated and together with the sequence of rewards $r_{t_n}$ are used to improve the agent over time.

Key in the OFENet structure are two networks implementing the functions $z_{o_t}=\phi_o(o_t)$ and $z_{o_t,a_t}=\phi_{o,a}(o_t,a_t)$ that take as inputs the observations (states) $o_t$ and actions $a_t$ available at time $t$ and produce the inflated feature vectors $z_{o_t}$ and $z_{o_t,a_t}$ as indicated. The latter are used in place of $o_t$ and $(o_t,a_t)$ in the networks of a particular RL agent such as SAC \cite{SAC_haarnoja2018soft}, TD3 \cite{TD3_fujimoto2018addressing} or PPO \cite{PPO_schulman_proximal_2017}. The networks $\phi_o$ and $\phi_{o,a}$ are implemented as DenseNets \cite{huang2017densely} and their structure is discussed more specifically in Section~\ref{sec:main}, where it is also modified to introduce pruning.  In the OFENet approach, the DenseNets $\phi_o$ and $\phi_{o,a}$ are trained concurrently with the training of the agent's RL network via an auxiliary task as illustrated in the lower right part of Figure~\ref{fig:oferl}. More specifically, an additional linear layer $\hat o_{t+1}=f_{pred}(z_{o_t,a_t})=W_{pred}\cdot z_{o_t,a_t}$ with weights $W_{pred}$ is trained along with $\phi_o$ and $\phi_{o,a}$ to predict the next observation $o_{t+1}$ by minimizing the cost function:
\begin{equation}\label{eq:OFE_pred}
	L_{pred} = \E_{\substack{o_t\sim p\\ a_t\sim\pi}}\norm{f_{pred}(z_{o_t,a_t})- o_{t+1}}^2\approx
\frac{1}{N-1}\sum_{n=1}^{N-1} \norm{f_{pred}(z_{o_{t_n},a_{t_n}})- o_{t_{n+1}}}^2
\end{equation}
where $o_{t_n}$, $a_{t_n}$ are available trajectory data stored in a buffer. In this sense, the combined network of $\phi_o$, $\phi_{o,a}$ and $f_{pred}$ serves as a model of the environment.
Subsequently, the trained networks $\phi$ and $\phi_{o,a}$ are used to produce the signals $z_{o_t}$ and $z_{o_t,a_t}$, respectively from the available signals $o_t$ and $a_t$. The former are then used as inputs to train RL networks such as the policy and value networks. Using the high-dimensional features $z_{o_t}$ and $z_{o_t,a_t}$ instead of the observation $o_t$ and action $a_t$ as inputs to several reinforcement learning algorithms such as SAC, TD3 or PPO has been shown to often increase performance \cite{ota2020can, ota_framework_2024}; however, these gains in performance, attributed to effectively enlarging the search space for the policy parameters \cite{ota2020can}, come at the cost of considerable increased complexity.

\subsection{Simultaneous training and pruning of Deep Reinforcement Learning Networks}\label{sec:RL}

A preliminary investigation of the application of  the pruning method of \cite{Guenter24_Robust, guenter2024complexityawaretrainingdeepneural} to the DNNs of RL algorithms is proposed in \cite[Ch.~6.5]{guenter_phdthesis} and is summarized in the following with additional details required for application to OFENets:
\begin{itemize}
	\item[1)] The NN models subject to simultaneous learning and pruning are  modified to stochastic XiNets by introducing Bernoulli RVs $\Xi$ at the output of each structure considered for elimination.
	Such models are typically fully-connected feed-forward neural networks and we simply implement this operation by following each hidden layer by a custom $\xi$-multiplication layer.
	For example, in the SAC algorithm \cite{SAC_haarnoja2018soft} enhanced with OFEnets the following DNNs are learned: the control policy network $\pi(a_t\mid z_{o_t})$, a value function network $V(z_{o_t})$, two Q-function networks $Q_i(z_{o_t,a_t}),\,i=1,2$,  as well as the OFE networks $\phi_o(o_t),\, \phi_{o,a}(o_t,a_t)$ and $W_{pred}$.
	
	\item[2)] Regularization costs on the NN weights $W$ (if the RL algorithm does not include them already) and the parameters $\Theta$  of the variational posterior distributions of the Bernoulli RVs serving  to promote sparsity are added to the maximization and/or regression objective functions of the RL algorithm and the auxiliary task of the OFE network. More specifically, let $\cal X$ denote the collection of the NN models employed by the specific RL algorithm (other than the OFENets), for example, the policy and value NNs. Then, for a model $x\in{\cal X}$, we add the following regularization cost to its training (minimization) objective:
	\begin{align}\label{eq:RL_reguloss}
		L_{x} = \frac{1}{2}\lambda_x\norm{W^x}^2
	+ \Gamma_x(\Theta^x)
	\end{align}
	with weight regularization hyperparameters $\lambda_x>0$.
	The complexity regularization term $\Gamma_x(\Theta^x)$ results from using the flattening hyperprior applied to each unit of the hidden layers of network ``$x$''  and if a common hyperparameter $0 <\gamma_x \leq 1$  is used for all units, it takes the form $\Gamma_x(\Theta^x)=-\log\gamma_x\cdot\|\Theta^x\|_1$ (compare to \eqref{eq:Cprn}).
	Alternatively, the hyperparameters of individual units can be automatically selected in terms of the parameters $\Theta^x$, so that $\Gamma_x(\Theta^x)$ expresses a scaled version of the expected computational/parameter complexity of the  $x$-network (see Section~\ref{sec:comp_cmpx}).

	\item[3)] During training, realizations  $\hat\Xi$ are sampled at each step that specify sub-networks for each RL network; the weights of these sub-networks are trained via backpropagation as usual. In addition to updating the NN's weight parameters $W$, the parameters $\Theta$ are updated via projected gradient descent and gradient approximations such as the Straight-Through estimator \cite{bengio2013estimating} (see \eqref{eq:theta_update} and \eqref{eq:deltaC}).

	\item[4)] During the simultaneous training and pruning process, the $\Theta$ parameters are monitored and once a $\theta$ parameter converges to $0$ (practically it becomes smaller that a specified threshold), the associated structure is permanently removed from the network either immediately or after a predefined number of additional gradient update steps.

\end{itemize}

\section{Main Results}\label{sec:main}

\subsection{Combining OFEnet and XiNet: the OFEXiNet}\label{sec:ofe-xi}
\begin{figure}[t!]
	\centering
	\includegraphics[width=1.0\textwidth]{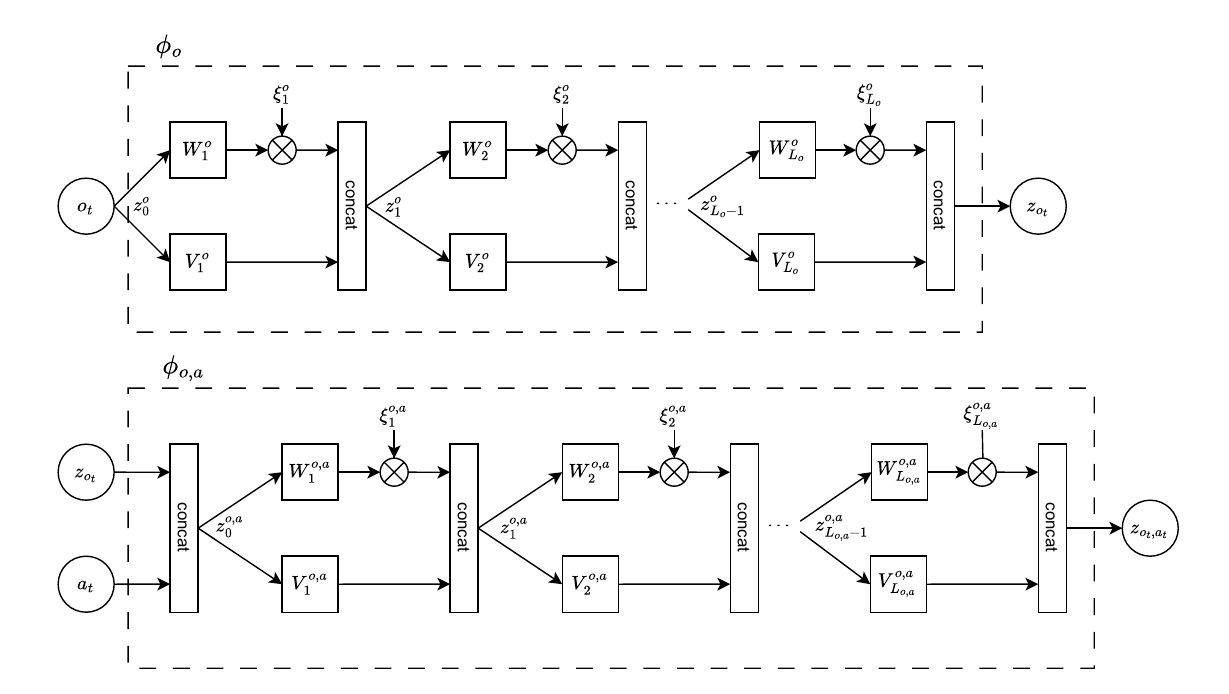}		
	\caption{The Online Feature Extractor (OFE) network with DenseNet \cite{huang2017densely} architecture and $\xi \sim Bernoulli(\theta)$ random variables. Each multiplication with $\xi$ is element-wise and ``concat'' denotes the concatenation of two vectors.}\label{fig:OFE}
\end{figure}

In this section, we discuss in detail how the OFE networks \cite{ota2020can} can be transformed to  XiNets for the purpose of simultaneous training and pruning. We accomplish this by introducing binary Bernoulli random variables $\Xi$ with learnable parameters $\Theta$ as outlined in Section~\ref{sec:RL}. The OFENets $\phi_o$ and $\phi_{o,a}$ are implemented as fully-connected DenseNets, which involves concatenating the outputs of the each hidden layer with its inputs to form the input to the next hidden layer. In our implementation, we insert a $\xi$-multiplication layer at the output of each hidden layer before it is concatenated with its input as shown in Figure~\ref{fig:OFE}. We also use additional learnable diagonal weights $V$ scaling the pass-through input before concatenation; we found that using the latter weights instead of the usual identity results in more efficient pruning.

The realized mappings by the two OFEXiNets $\phi_o$ and $\phi_{o,a}$ with $L_o$ and $L_{o,a}$ dense layer blocks, respectively can be expressed as follows.
First, $z_{o_t} = \phi_o(o_t;\Xi^o)$ is given by
\begin{align}\label{eq:struct_ofe_o}
\begin{split}
z_0^o &=o_t, \\
z_l^o&=\begin{bmatrix}
	\xi_l^o\odot a_l\left(BN(W_l^o z_{l-1}^o + b_l^o)\right)\\
	V_l^o z_{l-1}^o
\end{bmatrix}, \quad \xi_l^o\sim Bernoulli(\theta_l^o), \quad
\quad\text{for}\quad l=1 \dots L_o,\\
z_{o_t} &= z_{L_o}^o
\end{split}
\end{align}
and $z_{o_t, a_t} = \phi_{o,a}(z_{o_t}, a_t;\Xi^o,\Xi^{o,a})$ is given by
\begin{align}\label{eq:struct_ofe_oa}
	\begin{split}
		z_0^{o,a} &=\begin{bmatrix}
			z_{o_t}\\ a_t
		\end{bmatrix}, \\
		z_l^{o,a}&=\begin{bmatrix}
			\xi_l^{o,a}\odot a_l^{o,a}\left(BN(W_l^{o,a} z_{l-1}^{o,a} + b_l^{o,a})\right)\\
			V_l^{o,a} z_{l-1}^{o,a}
		\end{bmatrix}, \quad \xi_l^{o,a}\sim Bernoulli(\theta_l^{o,a}), \quad
		\quad\text{for}\quad l=1 \dots L_{o,a},\\
		z_{o_t, a_t} &= z_{L_{o,a}}^{o,a}.
	\end{split}
\end{align}
In these models, all  $\xi\sim Bernoulli(\theta)$ are discrete Bernoulli $0/1$ random vectors with a vector-valued parameter $\theta$ of appropriate dimension.
This means that the $i$th entry $[\xi]_i$ of a vector $\xi$ is Bernoulli distributed with parameter $[\theta]_i$, i.e., $[\xi]_i\sim Bernoulli\left([\theta]_i\right)$.
We collect all $\xi_l^o,\, l=1\dots L_o$ in $\Xi_o$ and all $\xi_l^{o,a},\, l=1\dots L_{o,a}$ in $\Xi_{o,a}$; similarly the $\theta$ parameters are collected in $\Theta^o$ and $\Theta^{o,a}$.
The weights $W_l^o, W_l^{o,a}$, biases $b_l^o, b_l^{o,a}$ and diagonal weights $V_l^o, V_l^{o,a}$ are of appropriate dimensions and collectively denoted as $M^o, M^{o,a}$.
The element-wise activation functions $a_l^o(\cdot), a_l^{o,a}(\cdot)$ are assumed to be twice continuously differentiable and satisfy $a_l(0)=0$.
Different realizations of the Bernoulli $\xi$ RVs implement sub-networks of structure~\eqref{eq:struct_ofe_o} and \eqref{eq:struct_ofe_oa}, where a unit of a fully-connected layer is either active or inactive, similar to unit-wise dropout. Similar to a Dropout  network \cite{dropout14srivastava}, the OFEXiNet can also be used  in ``evaluation mode'', where instead of sampling and multiplying units by $\Xi$, they are scaled by the parameters in $\Theta$.	
In this case, the resulting functions are denoted by $\phi_o(o_t;\Theta^o)$ and $\phi_{o,a}(z_{o_t},a_t;\Theta^o, \Theta^{o,a})$.
$BN(\cdot)$ denotes the use of a batch-normalization layer \cite{ioffe2015batch} for the purpose of combating covariance shift, smoothing the optimization landscape \cite{santurkar2018does} and improving learning; it is not explicitly shown in Figure~\ref{fig:OFE} and the remainder of this paper to simplify the notation.

The mappings $\phi_o$, $\phi_{o,a}$ are learned concurrently with the actual reinforcement learning task by minimizing the following auxiliary loss function comprising the loss of the auxiliary task $L_{pred}$ from \eqref{eq:OFE_pred} together with the OFENet regularization terms from {\eqref{eq:RL_reguloss}:
\begin{align}\label{eq:OFE_pred_reg}
\begin{split}
	L_{aux} = 	
	\frac{1}{N-1}\sum_{n=1}^{N-1} \bigg( \norm{f_{pred}(z_{o_{t_n},a_{t_n}})- o_{t_{n+1}}}^2\bigg)&
+ \frac{1}{2}\lambda_{OFE}\left(\norm{M^{o}}^2+\norm{M^{o,a}}^2\right)+\Gamma_{OFE}(\Theta^o, \Theta^{o,a})
\end{split}
\end{align}
where $z_{o_t,a_t}$ depends on $\Xi$.   In \eqref{eq:OFE_pred_reg},
$\lambda_{OFE}>0$ is the $\mathcal{L}_2$ weight regularization hyperparameter and $\Gamma_{OFE}(\Theta^o, \Theta^{o,a})$ represents regularization on the OFENet $\Theta$ parameters.
A general expression for $\Gamma_{OFE}$ is given by \eqref{eq:Cprn} resulting from the use of the flattening hyperprior and different hyperparameters $\gamma_k$ for the $k$th unit in the OFENets. When a common $\gamma_l$ is used for all units on the $l$th hidden layer, we obtain
\begin{align}\label{eq:OFE_cost}
\begin{split}
\Gamma_{OFE}(\Theta^o,\Theta^{o,a})=-\sum_{l=1}^{L_o}\log\gamma_l^o\cdot\|\theta_l^o\|_1
- \sum_{l=1}^{L_{o,a}}\log\gamma_l^{o,a}\cdot\|\theta_l^{o,a}\|_1.
\end{split}
\end{align}
In Section~\ref{sec:comp_cmpx}, we define $\log\gamma_i^o$ and $\log\gamma_i^{o,a}$ in terms of the parameters $\Theta^o$ and $\Theta^{o,a}$, respectively, so that $\Gamma_{OFE}$ becomes a scaled version of the expected computational/parameter complexity  introduced by the OFENets during the minimization of $L_{aux}$ and which turns out to be multilinear in $\Theta^o$, $\Theta^{o,a}$.

Solving
\begin{align}\label{eq:OFE_min}
	&\min_{M^o, M^{o,a}, \Theta^o, \Theta^{o,a}} L_{aux}\\
	&\quad s.t. \quad \Theta^o, \Theta^{o,a} \in \mathcal{H}
\end{align}
with
\begin{align}
	 &\mathcal{H}=\{\Theta=\{\theta_1,\dots,\theta_L\} \mid \forall\, l=1,\dots,L: \, \theta_l \in [0,1]^{I_l}\},	
\end{align}
where $I_l$ is the dimension of $\theta_l$, leads to a model that predicts higher-dimensional representations $z_{o_t,a_t}$ that allow to capture the dynamics of the system as a linear function of them. The data $(o_{t_n}, a_{t_n}, o_{t_n+1})$ required to approximate (see eq.\eqref{eq:OFE_pred}) and optimize $L_{aux}$ are sampled from an experience replay buffer $\mathcal{B}$, storing the trajectories visited by the reinforcement learning algorithm.
In addition, we estimate the expectation over the RVs in $\Xi$ with a single sample realization $\hat\Xi$ obtained during each forward pass. Then, the gradients with respect to the network weights $W$ are obtained by standard backpropagation. On the other hand, the gradients with respect to the $\Theta$ parameters are computed using equations \eqref{eq:theta_update} and \eqref{eq:deltaC} in Section~\ref{sec:pruning} with $L_{aux}$ from \eqref{eq:OFE_pred_reg} in place of $L$ in \eqref{eq:L_train_obj}.

When an entry of a $\theta$ parameter converges to $0$ during training, the corresponding unit becomes permanently inactive and can be removed from the network and further consideration.
In the supervised learning setting of \cite{guenter2024concurrenttraininglayerpruning, guenter2024complexityawaretrainingdeepneural} and with a multilinear regularization function $\Gamma_{OFE}$ in $\Theta^o$, $\Theta^{o,a}$, the optimal solution of \eqref{eq:OFE_min} is shown to be at one of extreme points of the hypercube $\mathcal{H}$, corresponding to a deterministic pruned network.
In the RL setting, where the distribution of the triplets ($o_t, a_t, o_{t+1}$) changes during the agent's learning, we still observe in our experiments that all $\theta$ values converge to either $0$ or $1$, albeit sometimes at a reduced rate.

\subsection{Sparsity Promoting Regularization and Computational Complexity}\label{sec:comp_cmpx}

In this section, we explicitly define the expected computational complexity cost $C_{OFE}(\Theta^o,\Theta^{o,a})$ for the OFEXiNets used in \eqref{eq:OFE_pred_reg} and also $C_x(\Theta^x)$ for the other RL networks. The factors $\log\gamma_i^o$ and $\log\gamma_i^{o,a}$ in \eqref{eq:OFE_cost} control the level of usefulness of the units in their corresponding layers
required for them to survive during the simultaneous training and pruning process. Thus, it is crucial for the effective reduction of OFENets that different such factors are used for each layer given their structure as DenseNets. Indeed, in the DenseNet Architecture copies of the outputs of units in layers closer to the input are used several times throughout the network; consequently, these units contribute much more to the computational and parameter complexity of the overall network. Therefore, it is imperative to set a higher threshold for the early layers to balance their contribution to cost reduction with their increased contribution to the network's complexity.

In particular, for the $\phi_o$ OFENet due to the fact that all weights $V_l^o$ are diagonal matrices, it can be easily seen from the concatenated intermediate features $z_l^o$ that every zero-entry in any $\xi_l^o,\, l=1,\dots,L_o$ leads to a zero-entry in the output $z_{o_t}=z_{L_o}^o$. Then, during simultaneous training and pruning when some $\theta$ parameter converges to $0$, the corresponding $\xi$ RV takes always a value of $0$ and the unit can be removed from the network without affecting it.
In this manner, the output dimension of $\phi_o(o_t)$ obtained after pruning depends on every layer's $\xi_l^o$. The same holds for the dimension of any intermediate $z_l^o$ in \eqref{eq:struct_ofe_o}, which affects the computation in each layer. As a result, operations in layers closer to the output become less expensive if any previous layer becomes narrower.
The same behavior is true for the $\phi_{o,a}$ OFENet since it employs a similar architecture to $\phi_{o}$.
In addition to the savings in the OFE network, since $z_{o_t}$ and $z_{o_t,a_t}$ are passed to the RL agent instead of the observation $o_t$ and the pair $(o_t, a_t)$, respectively, the operations of the agent's networks are more expensive if any layer in the OFE network remains wide.
In other words, the units of the earlier layers in the OFE network are re-used several times in the OFE network itself but also as the input of the RL agent and hence are more attractive to prune.

As discussed in Section~\ref{sec:pruning}, it is time-consuming to tune several $\log\gamma_k$ hyperparameters manually. Therefore, we employ here the approach of \cite{guenter2024complexityawaretrainingdeepneural}, which addresses the issue of cumbersome hyperparameter-tuning by defining a complexity-aware regularization cost {$\Gamma_{OFE}(\Theta)$ that represents the expected parameter complexity of the OFE network. To this end, we next
derive the expected computational complexity of the OFE networks formally in terms of the parameters in $\Theta$.

We focus on a single fully-connected layer with weight matrix $W_l^o$ (see Figure~\ref{fig:OFE}). We have previously established that the input vector $z_{l-1}^o$ to this matrix multiplication attains zero-entries according to all $\xi_1^o,\,\dots,\, \xi_{l-1}^o$. On the other hand, the output of this matrix multiplication and after passed through the activation function (not shown in Figure~\ref{fig:OFE}) is multiplied element-wise by $\xi_l^o$.
Then, for a given realization of the RVs $\hat\xi_i^o$, $i=1,\ldots,l$ and neglecting multiplications by the components of the vectors $\xi_i^o$ equal to $0$, the effective input and output dimensions of $W_l^o$ are $d_o + \sum_{i=1}^{l-1} \norm{\hat\xi_i^o}_1$ and
$\norm{\hat\xi_l^o}_1$, respectively, where $d_o$ is the dimensionality of an observation $o_t$.
Therefore, the computational load associated with this layer and sample $\hat\Xi$ can be expressed as
\begin{align}\label{eq:cost_Wo_xi}
   \left(d_o + \sum_{i=1}^{l-1} \norm{\xi_i^o}_1 \right) \cdot \norm{\xi_l^o}_1.
\end{align}
Next, we note that
\begin{align}
	\E_{\xi_i}\left[\norm{\xi_i}_1\right]= \norm{\theta_i}_1 \quad \text{and} \quad
\E_{\xi_i, \xi_j}\left[\norm{\xi_i}_1 \cdot  \norm{\xi_j}_1\right]= \norm{\theta_i}_1\cdot\norm{\theta_j}_1.
\end{align}
Then, taking expectations in \eqref{eq:cost_Wo_xi} leads to
\begin{align}\label{eq:cost_Wo}
   \left(d_o + \sum_{i=1}^{l-1} \norm{\theta_i^o}_1 \right) \cdot \norm{\theta_l^o}_1,
\end{align}
for the expected computational cost associated with layer $l$ of OFENet $\phi_o$. Clearly, \eqref{eq:cost_Wo} also gives the effective number of parameters in layer $l$.
Similarly, for any matrix multiplication $W_l^{o,a} z_{l-1}^{o,a}$ we obtain the expected computational load/effective number of parameters for layer $l$ as
\begin{align}\label{eq:cost_Woa}
	\left(d_o + \sum_{i=1}^{L_o} \norm{\theta_i^o}_1  + d_a + \sum_{i=1}^{l-1} \norm{\theta_i^{o,a}}_1 \right) \cdot \norm{\theta_l^{o,a}}_1,
\end{align}
where $d_a$ is the dimensionality of an action $a_t$.

Summing the terms in \eqref{eq:cost_Wo} and \eqref{eq:cost_Woa} over all layers and accounting for the operations involving adding the bias in each layer and the forward-propagation through BN layers leads to the expected computational complexity of $\phi_o(\cdot)$:
\begin{align}\label{eq:cmpx_o}
	C_o =\sum_{l=1}^{L_o} \left[ \left(d_o + \sum_{i=1}^{l-1} \norm{\theta_i^o}_1 \right) \cdot \left(1+ \norm{\theta_l^o}_1 \right) + 3\cdot\norm{\theta_l^o}_1\right].
\end{align}
and of $\phi_{o,a}(\cdot)$:
\begin{align}\label{eq:cmpx_o_a}
	C_{o,a} = \sum_{l=1}^{L_{o,a}} \left[ \left( d_o + \sum_{i=1}^{L_o} \norm{\theta_i^o}_1 + d_a + \sum_{i=1}^{l-1}\norm{\theta_i^{o,a}}_1 \right) \cdot \left(1+ \norm{\theta_l^{o,a}}_1 \right) + 3\cdot \norm{\theta_l^{o,a}}_1 \right]
	\end{align}
in the case of fully connected layers.
In the above equations the ``$+3$''-term in eq. \eqref{eq:cmpx_o} and \eqref{eq:cmpx_o_a} reflects the computation of the bias and the batch-normalization constants. Note that this constant can be increased to account for the computation of the nonlinear activation function. Also, the ``$1+$'' in eq. \eqref{eq:cmpx_o} and \eqref{eq:cmpx_o_a} accounts for the computation of the diagonal weights $V$. We also remark that this formulation can be easily extended to convolutional layers, as done in \cite{guenter2024complexityawaretrainingdeepneural} for residual network (ResNet \cite{he2016deep}) architectures \cite{he2016deep}.

An additional complexity associated with the OFENets results from the computation of $f_{pred}=W_{pred}\cdot z_{o_t,a_t}$ during their training and from the fact that their output becomes the input for several of the RL networks in $\cal X$, such as the policy and value NNs. More specifically, the computational complexity to predict the next observation through the function $f_{pred}(\cdot)$ is
\begin{align}
	C_{pred} = \left(1+d_o+d_a+\sum_{i=1}^{L_o} \norm{\theta_i^o}_1 + \sum_{i=1}^{L_{o,a}} \norm{\theta_i^{o,a}}_1 \right) \cdot d_o.
\end{align}
Then, the computational complexity of the first stage of network ``$x$'' receiving input from an OFENet of dimension $D_i^x(\Theta^o, \Theta^{o,a})$  and first layer with corresponding vector of $\theta$ parameters given by $\theta^x_1$ is expressed as
\begin{align}\label{eq:OFEx}
	C_{x,OFE} = \left(1+D_i^x(\Theta^o, \Theta^{o,a})\right) \cdot \norm{\theta^x_1}_1.
\end{align}
In \eqref{eq:OFEx}, it is
\begin{align}\label{eq:dxo}
	D_i^x(\Theta^o) =  d_o + \sum_{l=1}^{L_o} \norm{\theta_l^o}_1,
\end{align}
or
\begin{align}\label{eq:dxoa}
	D_i^x(\Theta^o,\Theta^{o,a}) =  d_o + \sum_{l=1}^{L_o} \norm{\theta_l^o}_1 + d_a + \sum_{l=1}^{L_{o,a}}\norm{\theta_l^{o,a}}_1.
\end{align}
if the network ``$x$'' takes $z_{o_t}$ or $z_{o_t,a_t}$ as input, respectively.

It is important to note that from the two OFENets $\phi_{o}$ and $\phi_{o,a}$, only the first is needed for deployment of a final policy on a device that is not subject to further learning. Table~\ref{tab:model_usage} summarizes this property for the several possible NN models in a RL algorithm with OFENets.
Depending on whether the goal is to perpetually learn and try to improve the control policy or to find a near optimal control policy in limited time, we define the weighted computational cost of the OFENets using a discount factor $0\leq\rho\leq 1$ as follows:
\begin{align}\label{eq:C-match}
C_{OFE}=C_o + C_{\pi, OFE} + \rho \left(C_{o,a}+C_{pred} + \sum_{x\in \mathcal{X}\setminus\pi}C_{x,OFE}\right),
\end{align}
where $\pi\in{\cal X}$ denotes the policy network.
Then, depending on the situation at hand, one may choose either $\rho=0$ if the reduction of network complexity for deployment only is the main concern, or $\rho=1$ if that for continual learning is the focus. Choosing an intermediate $0<\rho<1$ balances the two extremes.

Next, let $\mathcal{X}_o$ and $\mathcal{X}_{o,a}$ be the groups of RL networks that take $z_{o_t}$ and $z_{o_t,a_t}$, respectively as inputs and $\mathcal{X} =\mathcal{X}_o \cup \mathcal{X}_{o,a}$; in particular, the control policy $\pi \in\mathcal{X}_o$.
By taking in \eqref{eq:OFE_cost} $\log\gamma_l^o$ and $\log\gamma_l^{o,a}$ to satisfy
for $l=1,\dots,L_o$:
\begin{align}\label{eq:gamma_faco}
	\log\gamma_l^o =-\nu_{OFE}\cdot\left(
	(1+\rho)\cdot d_o + \sum_{i=1}^{l-1}\|\theta_i^o\|_1+L_o-l +\rho L_{o,a} + 3  +\|\theta_1^\pi\|_1+ \rho\sum_{x\in{\mathcal{X}_o \setminus\pi}}\|\theta_1^x\|_1
	\right)
\end{align}
and for $l=1,\dots, L_{o,a}$:
\begin{align}\label{eq:gamma_facoa}
	\log\gamma_l^{o,a} =-\nu_{OFE}\cdot\rho\cdot\left(
	2d_o+d_a +\sum_{i=1}^{L_o}\|\theta_i^o\|_1+ \sum_{i=1}^{l-1}\|\theta_i^{o,a}\|_1
	+L_{o,a}-l+3+\sum_{x\in{\cal X}_{o,a}}\|\theta_1^x\|_1
	\right)
\end{align}
we obtain
\begin{align}\label{eq:C-match}
\Gamma_{OFE}=\nu_{OFE}\cdot C_{OFE} + const.
\end{align}
Thus, the regularization term $\Gamma_{OFE}(\Theta^o, \Theta^{0,a})$ in \eqref{eq:OFE_pred_reg} matches the OFENet complexity $C_{OFE}$  within a scaling factor $\nu_{OFE}>0$ that can be used to trade-off performance vs. complexity and a constant that can be ignored for the optimization in \eqref{eq:OFE_min}.

Next, we state the computational complexity of a fully-connected feed-forward NN as used in many RL algorithms for the control policy and value function networks.
For a network $x\in{\cal X}$ with $L_x$ layers and parameters $W^x$ and $\Theta^x$, we obtain
\begin{align}
	C_x = \left(1+D_i^x(\Theta^o, \Theta^{o,a})\right) \cdot \norm{\theta^x_1}_1 + \sum_{l=1}^{L_x-1}\bigg((1+\norm{\theta_l^x}_1)\cdot\norm{\theta_{l+1}^x}_1\bigg) + \left(1+\norm{\theta^x_{L_x}}_1\right)\cdot D_o^x
\end{align}
with $C_x$ depending on its parameters $\Theta^x$, its constant output dimension $D_o^x$ and its input dimension $D_i^x(\Theta^o,\Theta^{o,a})$, a function of the output dimension of the OFE network providing its input (see \eqref{eq:dxo} and \eqref{eq:dxoa}) and hence a function of all $\theta$ parameters in the OFE network.

\begin{table}[!t]
	\centering
	\caption{Overview over the different NN models at the example of a SAC RL agent \cite{SAC_haarnoja2018soft} and whether they are used for learning only or also during deployment of a final control policy.}\label{tab:model_usage}
	\begin{tabular}{l|l|l}
		Model & Learning & Deployment  \\
		\hline
		OFENet $\phi_{o}(o_t)$     & \cmark       & \cmark         \\
		OFENet $\phi_{o,a}(o_t,a_t)$    & \cmark       & \xmark          \\
		State predictor $f_{pred}(z_{o_t,a_t})$  & \cmark       & \xmark           \\
		Policy $\pi(a_t\mid z_{o_t})$ & \cmark    & \cmark           \\
		Value function $V(z_{o_t})$     & \cmark        & \xmark           \\
		Q function $Q(z_{o_t,a_t})$     & \cmark        & \xmark
	\end{tabular}
\end{table}

As with the OFENets, we can select the flattening hyperparameters $\log\gamma_l^x$ for the units of network $x\in\cal X$ from
\begin{align}
	\begin{split}
		\log\gamma_1^x&=-\nu_x\left(1+D_i^x\right), \quad\quad \log\gamma_{L_x} = -\nu_x\left( D_o^x + 1 +\norm{\theta_{L_x-1}^x}_1\right)\\
			\text{and}\quad\log\gamma_l^x &= -\nu_x\left(1+\norm{\theta_{l-1}^x}_1\right)\quad \text{for} \quad l=2,\dots,L_x-1,.
	\end{split}
\end{align}
so that  $\Gamma_x$ matches its computational complexity within a scaling factor $\nu_x>0$, namely $\Gamma_x=\nu_x\cdot C_x+const.$

In summary, with the choice of $\log\gamma_l^o,\, \log\gamma_l^{o,a}$ from \eqref{eq:gamma_faco}, \eqref{eq:gamma_facoa}, respectively, we train the OFENets by minimizing
\begin{align}\label{eq:OFE_pred_reg_cmpx}
	\begin{split}
		L_{aux} = 	
		\frac{1}{N-1}\sum_{n=1}^{N-1} \left( \norm{f_{pred}(z_{o_{t_n},a_{t_n}})- o_{t_{n+1}}}^2\right)
 +\frac{1}{2}\lambda_{OFE}\left(\norm{M^{o}}^2+\norm{M^{o,a}}^2\right)
 +\nu_{OFE}\cdot C_{OFE}
	\end{split}
\end{align}
instead of \eqref{eq:OFE_pred_reg} and  we add the regularization loss
\begin{align}\label{eq:RL_reguloss_cmpx}
	L_{x} = \frac{1}{2}\lambda_x\norm{W^x}^2
		+\nu_x\cdot C_x
\end{align}
to the corresponding objective functions of all RL networks instead that of \eqref{eq:RL_reguloss}.

\subsection{Proposed Algorithm}\label{sec:algo}
Algorithm~\ref{alg1} summarizes the training of the OFEXiNet, which occurs concurrently with learning the RL agent.
The algorithm requires as inputs the the agent's NNs models $\mathcal{X}$, learning rates $\eta_{OFE}>0$ for the OFEXiNet and $\eta_{RL}>0$ for the Xi-networks in $\mathcal{X}$, and regularization parameters $\nu_{OFE}>0$, $\nu_{RL}=\{\nu_x>0\}_{x\in\mathcal{X}}$ and $\lambda_{OFE},\,\lambda_{RL}>0,$ for the sparsity promoting regularization losses in \eqref{eq:OFE_pred_reg_cmpx} and \eqref{eq:RL_reguloss_cmpx}.
Further, initial parameters of the OFEXiNet: $M_{OFE}=\{W^{o}, W^{o,a}, W^{pred}, V^{o}, V^{o,a} \},\, \Theta_{OFE}=\{ \Theta^o,\Theta^{o,a}\}$ as well as initial parameters of the RL algorithm: $M_{RL}= \{W^x\}_{x\in\mathcal{X}},\, \Theta_{\cal X}=\{\Theta^x\}_{x\in \cal X}$ are needed. Lastly, the experience replay buffer: $\mathcal{B}=\{\}$ is initialized and a small pruning tolerance: $\theta_{tol}>0$ is selected.

The proposed algorithm consists of four main parts. First, the current OFENet $\phi_o(o_t;\Theta^o_t)$ and control policy $\pi(z_t;\Theta^\pi_t)$ are used to select the action $a_t$, which is subsequently applied to the environment, resulting in the observation $o_{t+1}$ and reward $r_t$. Note that in this step the OFEXiNet and the policy network
are used in ``evaluation mode'', i.e., the corresponding parameters $\Theta$ are used instead of samples $\hat\Xi$ to avoid introducing additional noise in the selection of $a_t$. The obtained data is added to the replay buffer $\mathcal{B}$.

In the second part of the algorithm, the OFEXiNets are updated by first sampling a triplet \sloppy $\{o_{t,\mathcal{B}}, a_{t,\mathcal{B}}, o_{t+1, \mathcal{B}}\}$ from $\mathcal{B}$ and obtaining samples $\hat \Xi^o, \hat \Xi^{o,a}$ for the training of the OFEXiNets. Next, we estimate $L_{aux}$ from \eqref{eq:OFE_pred_reg_cmpx} using these samples and take a gradient descent step on all $W,V$ parameters in $M_{OFE}$ of the OFE sub-networks defined by $\hat \Xi^o, \hat \Xi^{o,a}$ and a projected gradient descent step on the $\Theta^o$, $\Theta^{o,a}$ parameters.
To calculate the gradients with respect to $\Theta^o$ and $\Theta^{o,a}$, we use \eqref{eq:theta_update} and \eqref{eq:deltaC} with $L_{aux}$ replacing $L$.
Next, new data $\{o_{t,\mathcal{B}}, a_{t,\mathcal{B}}, o_{t+1, \mathcal{B}}\}$ is sampled from $\mathcal{B}$ and
the features $z_o=\phi_{o}(o_{t,\mathcal{B}};\Theta^o)$ as well as $z_{o,a}=\phi_{o,a}(z_o,a_{t,\mathcal{B}};\Theta^o, \Theta^{o,a})$ are calculated using the OFENet in evaluation mode.

The third part of the algorithm consists of the RL agent update. As described in Section~\ref{sec:RL}, eq. \eqref{eq:RL_reguloss_cmpx} is added for each model in $\mathcal{X}$ to its corresponding loss function. Due to the usage of the OFEXiNet, we replace $o_t$ and $(o_t, a_t)$ with the previously obtained $z_{o}$ and $z_{o,a}$, respectively as inputs to the RL agent.
For each RL network updated, a realization of the involved $\Xi$ variables is sampled and the weights of the resulting sub-network are learned as specified by the RL algorithm; in addition, the $\Theta$ parameters of the trained RL network are updated using projected gradient descent based on corresponding equations to \eqref{eq:theta_update} and the straight-through approximation \eqref{eq:deltaC}. Any RL networks not updated at this time are used in ``evaluation mode''; for example, in SAC \cite{SAC_haarnoja2018soft},  when calculating the gradients for the policy network, we use the Q-function networks in evaluation mode and vice versa.

During the fourth part of the algorithm structures of the network multiplied by $\xi$ RVs with corresponding $\theta$ parameters that converge to zero are removed. These structures are inactive most of the time and do not contribute to the networks output. Specifically, we prune any unit with $\theta < \theta_{tol}$, a small fixed pruning tolerance, e.g., $\theta_{tol}=0.1$.
If the algorithm has not converged or a fixed number of environment steps is reached, we repeat all parts of the algorithm.

\begin{algorithm}[!t]
	\caption{Training/Pruning of OFENetPrune}
	\label{alg1}	
	\begin{algorithmic}[1]
		\REQUIRE Set of the agent's NNs: $\mathcal{X}$, learning rates: $\eta_{OFE}, \,\eta_{RL}>0$, regularization parameters:
		$\nu_{OFE}>0, \nu_{RL}=\{\nu_x>0\}_{x\in\mathcal{X}}$
 		and $\lambda_{OFE}>0, \lambda_{RL}=\{\lambda_x>0\}_{x\in\mathcal{X}}$, initial parameters of OFEXiNet: $M_{OFE}=\{W^{o}, W^{o,a}, W^{pred}, V^{o}, V^{o,a}\},\,\Theta_{OFE}=\{\Theta^{o}, \Theta^{o,a}\}$,
		initial parameters of the RL algorithm: $M_{RL}= \{W^x\}_{x\in\mathcal{X}},\, \Theta_{\cal X}=\{\Theta^x\}_{x\in \cal X}$,
		experience replay buffer: $\mathcal{B}=\{\}$, pruning tolerance: $\theta_{tol}>0$.
		
		\FOR{each environment step}
		\vspace{0.2cm} \hrule \vspace{0.2cm}
		\STATE\textbf{I: Sampling Trajectories}
		\STATE $a_t \sim \pi\left(a_t | \phi_o(o_t;\Theta^o);\Theta^\pi\right)$
		\STATE $o_{t+1} \sim p(o_{t+1}|o_t, a_t)$
		\STATE $\mathcal{B} \leftarrow \mathcal{B} \cup \{(o_t, a_t, o_{t+1}, r_{t+1})\}$
		\vspace{0.2cm} \hrule \vspace{0.2cm}
		\STATE\textbf{II: OFEXiNet Update}
		\STATE sample mini-batch $\{o_{t,\mathcal{B}}, a_{t,\mathcal{B}}, o_{t+1, \mathcal{B}}\}$ from $\mathcal{B}$
		\STATE sample $\hat \Xi^o$, $\hat \Xi^{o,a}$ and estimate $L_{aux}$ from \eqref{eq:OFE_pred_reg_cmpx}
		\STATE take one step $\begin{bmatrix}M_{OFE} & \Theta_{OFE}\end{bmatrix} \leftarrow \begin{bmatrix}M_{OFE} & \Theta_{OFE}\end{bmatrix} - \eta_{OFE} \nabla_{\tiny\begin{bmatrix}M_{OFE} & \Theta_{OFE}\end{bmatrix}} L_{aux}$
        \STATE project $\Theta_{OFE}$ in $[0,\, 1]$
		\STATE resample mini-batch $\{o_{t,\mathcal{B}}, a_{t,\mathcal{B}}, o_{t+1, \mathcal{B}}\}$ from $\mathcal{B}$
		\STATE $z_{o} \leftarrow \phi_o(o_{t,\mathcal{B}};\Theta^o)$ \label{line:zo}
		\STATE $z_{o,a} \leftarrow \phi_{o,a}(z_o, a_{t,\mathcal{B}};\Theta^o,\Theta^{o,a})$ \label{line:zoa}
		\vspace{0.2cm} \hrule \vspace{0.2cm}
		\STATE\textbf{III: RL Agent Update}
		\STATE Add $L_x$ from \eqref{eq:RL_reguloss_cmpx} for each model in $\mathcal{X}$ to its corresponding loss function
		\STATE Update the agent parameters $M_{RL},\, \Theta_{\cal X}$ using learning rate $n_{RL}$ and the representations $z_{o}$ and $z_{o,a}$ from line \ref{line:zo} and \ref{line:zoa} instead of $o_t$ and $(o_t,\, a_t)$, respectively
		\STATE project $\Theta_{\cal X}$ in $[0,\, 1]$
		\vspace{0.2cm} \hrule \vspace{0.2cm}
		\STATE\textbf{IV: Pruning Phase}
		\FOR{each $\theta$ in $\Theta_{OFE}$ and $\Theta_{\cal X}$}
		\IF{an entry $[\theta]_i < \theta_{tol}$}
		\STATE prune the unit and its corresponding weights from the network
		\ENDIF
		\ENDFOR
		\vspace{0.2cm} \hrule \vspace{0.2cm}
		\ENDFOR
	\end{algorithmic}
\end{algorithm}

\section{Related Work}\label{sec:relwork}

\noindent\emph{Training Instabilities and Overparametrization in RL:}
\cite{song2019observational, zhang2018studyoverfitting} study the problem of overfitting in deep RL and find that standard RL agents may overfit in various ways, ultimately resulting in reduced generalization and performance in unseen environments. \cite{henderson2018deep} compares the performance on deep RL algorithms with using networks of different number of units and layers, and
finds that larger networks do not consistently improve performance.

A variety of regularization and stabilization techniques have been proposed to address these issues.
For example, \cite{gal2016dropout} uses Dropout and Bayesian NNs to improve robustness in RL. \cite{hiraoka2022dropout} uses in addition layer normalization in their NN models within advanced actor-critic algorithms. Lastly, \cite{liu2020regularization} shows that $\mathcal{L}_2$ weight regularization can enhance both on-policy and off-policy RL algorithms.
\newline
\newline
\noindent\emph{Sparse Training and Pruning in RL:}
While scaling up neural architectures has become a common approach in supervised learning, recent work highlights unique challenges regarding parameterization and model efficiency in RL. Notably, \cite{kumar_implicit_2021} identifies an implicit under-parametrization phenomenon in value-based RL methods. More gradient updates lead to a decrease in expressivity of the current value NN, characterized via a drop in the rank of the value network's features; it is shown in simulations that this typically leads to a performance loss. This insight contrasts with the supervised learning paradigm, where over-parameterized models are typically favored for their expressive capacity.

A growing body of literature is exploring the potential of sparse training methods as a remedy to over- or under-parameterization in RL. \cite{graesser2022state} provides a comprehensive study on the state of sparse training in deep RL showing that performance can be maintained while reducing the model to a small fraction of its original parameter count; the provided experiments across algorithms such as DQN \cite{mnih_playing_2013}, PPO, and SAC attest the efficacy of such approaches.

\cite{Grooten23_filtering} proposes an automatic noise filtering method that continually adjusts network connectivity to focus on task-relevant features. Applied to algorithms like SAC and TD3, this method dynamically refines the network structure during training, enhancing both efficiency and performance.

Value-based agents can also benefit substantially from structured pruning approaches. \cite{Ceron24_invaluebased} demonstrates that gradual magnitude pruning allows agents to maximize the effectiveness of their parameters; such results indicate that networks pruned in this manner can achieve dramatic performance improvements using only a small fraction of the full set of network parameters.

The potential of simultaneous training and pruning \cite{guenter2024complexityawaretrainingdeepneural, Guenter24_Robust} (see Section~\ref{sec:pruning}) for addressing the unique challenges of RL, such as instability, over-fitting, and the high computational demands have been explored in \cite{guenter_phdthesis}, where it is applied to RL methods, such as SAC \cite{SAC_haarnoja2018soft}, TD3 \cite{TD3_fujimoto2018addressing} and PPO \cite{PPO_schulman_proximal_2017}. In this line of work, agent networks are first replaced by scaled, residual-type architectures prior to pruning.
\newline
\newline
\noindent\emph{Scaling RL Agent's Networks:}
Recent advances have demonstrated that, with suitable modifications, scaling neural network architectures in RL can yield significant performance improvements.

\cite{schwarzer_bff} uses a scalable ResNet architecture together with large replay ratios and periodic resets of network layers to achieve a breakthrough in model-free RL by attaining human-level performance and sample efficiency on typical benchmarks.

\cite{munk2016learning} first uses representation learning for the purpose of RL. A model network is learned and is subsequently used to pre-train the first layer of the actor and critic networks. In this work, the goal is to learn a compact representation from the noisy and high dimensional observations of the environment and it is shown that the method can significantly improve the final performance in comparison to end-to-end learning.
Building on the insight that representation quality is crucial in RL, \cite{protovaluenet_farebrother2023} proposes ``Proto-Value Networks'', leveraging auxiliary tasks to enhance representation learning. These results show that expanding both the number of auxiliary tasks and the network size consistently leads to better performance, suggesting that scalable architectures, coupled with rich training signals, can be particularly beneficial.
\cite{ota2020can} introduces the OFE network which we have discussed in Section~\ref{sec:OFE}.

\section{Simulation Experiments}\label{sec:simulations}

We evaluate our method using a Soft Actor Critic (SAC) \cite{SAC_haarnoja2018soft} agent on the popular robotic control environments of the \text{Gymnasium} \cite{gymnasium2024towers} suite with MuJoCo. In particular, the HalfCheetah-v2, Ant-v2, Hopper-v2 and Walker2D-v2 environments are considered.
To demonstrate the effectiveness of our approach we compare it to multiple existing methods listed next:
\begin{description}
	\item[SAC:] the standard SAC algorithm \cite{SAC_haarnoja2018soft} using networks with 256 units in each of the two hidden-layers.
	\item[SAC-Big:]  the standard SAC algorithm \cite{SAC_haarnoja2018soft} using networks with 1024 units in each of the two hidden-layers.
	\item[OFE:]  the standard OFE algorithm \cite{ota2020can} with a SAC agent using 32 units per OFE-network layer and 256 units in the 2-hidden-layer SAC networks.
	\item[OFE-Big:]  the standard OFE algorithm \cite{ota2020can} with a SAC agent using 128 units per OFE-network layer and 512 units in the 2-hidden-layer SAC networks.
	\item[Prune:]  our method, that prunes both OFE networks and all networks of its SAC agent. Starting at 128 units per OFE layer and 512 units in the SAC networks. We run our method with three sets of hyperparameters to achieve various trafe-offs between NN reduction rates and performance levels. These parameter choices were not tuned to perform best on a single environment and rather reflect how their selection influences the behavior of our method. The different selections are denoted by Prune-A, Prune-B and Prune-C in the remainder of this section.
\end{description}

Using SAC as the agent, several DNNs are learned in our method: the control policy network $\pi(a_t\mid z_{o_t};W^\pi, \Theta^\pi)$, a value function network $V(z_{o_t};W^v, \Theta^v)$,  two Q-function networks $Q_1(z_{o_t,a_t};W^{q_1},\Theta^{q_1})$, $Q_2(z_{o_t,a_t};W^{q_2}, \Theta^{q_2})$, as well as the OFE networks $\phi_o(o_t;W^o, V^o,\Theta^o)$, $\phi_{o,a}(o_t,a_t;W^{o,a}, V^{o,a}, \Theta^{o,a})$  and the prediction network $f_{pred}(z_{o,a};W_{pred})$. Therefore, for the SAC agent, we have in an obvious notation $\mathcal{X}=\{\pi,v,q_1,q_2\}$.
Similar to \cite{ota2020can}, we use OFENets with $L_o=L_{o,a}=6$ layers for the Ant, Hopper and Walker environments and $L_o=L_{o,a}=8$ layers for the Cheetah environment.

We use the Adam optimizer \cite{kingma2014adam} with learning rates of $\eta_{OFE}=\expnumber{3}{-4}$ and $\eta_{RL}=\expnumber{3}{-4}$.
The values of the regularization hyperparameters used in our Prune-A, Prune-B and Prune-C experiments for the OFENets and the RL networks in $\mathcal{X}=\{\pi,v,q_1,q_2\}$ are listed in Table~\ref{tab:prune_hyperparams}; we also report there the value of the parameter $\rho$ introduced in Section~\ref{sec:comp_cmpx} to trade-off complexity during training and deployment.
All $\theta$ parameters are initialized at $1$ and fixed for the first $200,000$ steps. For the last $20\%$ of training, all $\theta$ parameters that have not yet completely converged to either $0$ or $1$ are rounded and fixed at $0$ or $1$, so that at the end of training a deterministic network results. We train each agent for $\expnumber{2}{6}$ environment steps in the case of the Cheetah environment and for $\expnumber{1}{6}$ steps in the case of all other environments.
Similar to \cite{ota2020can} and with the goal to eliminate dependency on the specific initialization
of the policy, we use a random policy to fill the replay buffer for the first $\expnumber{1}{4}$ time steps.
We also adopt the short pre-training phase of OFENet on these samples from \cite{ota2020can}.

\begin{table}[t!]
	\renewcommand{\arraystretch}{1.2} 
	\centering
	\begin{tabular}{l|c|c|c}
		{Hyperparameter}     & \textbf{Prune-A} & \textbf{Prune-B} & \textbf{Prune-C} \\
		\hline
		$\lambda_{OFE}$      &     $\expnumber{1}{-9}$              &     $\expnumber{1}{-6}   $          &      $\expnumber{1}{-6} $             \\
		$\lambda_{RL}$      &       $\expnumber{1}{-9}$            &          $\expnumber{1}{-9}  $        &      $\expnumber{1}{-9}$             \\
		\hline
		$\nu_{OFE}$        &     $\expnumber{5}{-8}$               &     $\expnumber{2}{-7}$               &    $\expnumber{3}{-7}$                \\
		$\nu_{\pi}$        &    $\expnumber{2}{-6}$                &     $\expnumber{1}{-5}$               &   $\expnumber{2}{-5}$                 \\
		$\nu_{v}$        &         $\expnumber{1}{-4}$           &     $ \expnumber{5}{-4}$               &     $\expnumber{2.5}{-3}$               \\
		$\nu_{q_i},\, i=1,2$        &      $\expnumber{1}{-4}$              &        $ \expnumber{5}{-4}$            &   $\expnumber{2.5}{-3}$                 \\
		\hline
		$\rho$                       &      $  1.0  $        &    $  0.5 $           &      $ 0.5   $        \\
	\end{tabular}
	\caption{Regularization hyperparameters of \eqref{eq:OFE_pred_reg_cmpx} and \eqref{eq:RL_reguloss_cmpx} used in different pruning experiments with our method: Prune-A, Prune-B and Prune-C.}
	\label{tab:prune_hyperparams}
\end{table}

Table~\ref{tab:scores_PR} shows, in the top row of each cell, the highest score (cumulative reward) seen when evaluating the current control policy during the last 20\% of training. In this period, our XiNets have already fixed $\theta$ values at either $0$ or $1$ and hence have attained their final sizes and are deterministic. Below each score we report the parameter deploy ratio (dR) on the left and the parameter train ratio (tR) on the right with respect to the OFE baseline method.
These ratios describe the fraction of the total number of parameters remaining in the groups of networks for deployment and training (see Table~\ref{tab:model_usage}) with respect to the OFE method baseline. A percentage less than 100\% indicates that the collection of networks used for deployment/training are smaller than those in the baseline method.
We note that since we use fully-connected NNs, the parameter complexity is closely related to the computational complexity, since no parameters are re-used unlike, for example, in the case of convolutional NNs.
Our results show that by sizing up SAC to SAC-Big higher scores are consistently achieved at the expense of about 15 times the computational/parameter complexity. Similarly OFE-Big outperforms OFE in terms of the accumulated rewards at the expense of a about 6-7 times higher complexity.

In the case of the Hopper environment our method with Prune-A yields much smaller networks for deployment and training when compared to OFE method while maintaining a higher score. For the other three environments our method, starting from the network architecture of the OFE-Big method, prunes the number parameters to at least half while maintaining at least 95\% of the OFE-Big baseline score. The scores of Prune-A are higher than the ones obtained with the OFE baseline at the expense of larger networks.
Prune-B maintains at least 94\%, 96\% and 99\% of the score of the OFE method in case of the  Walker2D, Cheetah and Hopper environments respectively. This is achieved with much smaller deployment networks ($<50\%$ dR) and also much smaller ($<50\%$ tR) networks for training in case of the Cheetah and Hopper environments.
Prune-B shows also that our method is able to find, higher performing but smaller control policies (low dR) when compared to the standard SAC algorithm. In all four environments lower dR are achieved at a much better score when compared to SAC.
Prune-C shows that when increasing the regularization, our method also achieves small tR while maintaining a score higher than that of SAC and often close to that of the much larger SAC-Big networks.

\begin{table}[h]
	\renewcommand{\arraystretch}{2.5} 
	\centering
	\caption{Each cell shows the score (cumulative reward during evaluation) in its first row of our method and several existing methods. The deploy ratios (dR) and train ratios (tR) of the network parameters are shown below each score on the left and right, respectively. A percentage smaller than 100\% indicates that the resulting networks are smaller than the baseline networks given by the OFE method.}
	\label{tab:scores_PR}
	\begin{tabular}{l|c|c|c|c}
		\makecell{Score\\ \small dR \,\, tR}& \textbf{HalfCheetah-v2} & \textbf{Ant-v2} & \textbf{Hopper-v2} & \textbf{Walker2D-v2} \\
		\hline
		SAC & \makecell{12757 \\
			\small 42\%\,\, 35\%} &
		\makecell{4974 \\
			\small 53\%\,\, 49\%} &
		\makecell{3197 \\
			\small 51\%\,\, 43\%} &
		\makecell{3998 \\
			\small 52\%\,\, 44\%} \\
		\hline
		SAC-Big & \makecell{15534 \\
			\small 621\%\,\, 524\%} &
		\makecell{6345 \\
			\small 635\%\,\, 588\%} &
		\makecell{3534 \\
			\small 773\%\,\, 656\%} &
		\makecell{4976 \\
			\small 759\%\,\, 649\%} \\
		\hline
		OFE & \makecell{16609 \\
			\small 100\%\,\, 100\%} &
		\makecell{7161 \\
			\small 100\%\,\, 100\%} &
		\makecell{3548 \\
			\small 100\%\,\, 100\%} &
		\makecell{5468 \\
			\small 100\%\,\, 100\%} \\
		\hline
		OFE-Big & \makecell{17299 \\
			\small 739\%\,\, 765\%} &
		\makecell{7608 \\
			\small 565\%\,\, 640\%} &
		\makecell{3699 \\
			\small 665\%\,\, 702\%} &
		\makecell{5490 \\
			\small 653\%\,\, 696\%} \\
		\hline
		\hline
		Prune-A & \makecell{16774 \\
			\small 87\%\,\, 179\%} &
		\makecell{7229 \\
			\small 130\%\,\, 139\%} &
		\makecell{3659 \\
			\small 39\%\,\, 86\%} &
		\makecell{5679 \\
			\small 253\%\,\, 330\%} \\
		\hline
		Prune-B & \makecell{16033 \\
			\small 16\%\,\, 47\%} &
		\makecell{6517 \\
			\small 39\%\,\, 72\%} &
		\makecell{3515 \\
			\small 10\%\,\, 38\%} &
		\makecell{5152 \\
			\small 47\%\,\, 88\%} \\
		\hline
		Prune-C & \makecell{13593 \\
			\small 12\%\,\, 29\%} &
		\makecell{5829 \\
			\small 22\%\,\, 33\%} &
		\makecell{3470 \\
			\small 6\%\,\, 16\%} &
		\makecell{4803 \\
			\small 35\%\,\, 59\%} \\
	\end{tabular}
\end{table}

Table~\ref{tab:cmpx_new} shows the number of parameters for deployment (d) of the control policy in each cell on the left and the number of parameters for continued learning (t) on the right. Both quantities are in thousands.

\begin{table}[!h]
	\renewcommand{\arraystretch}{1.25} 
	\centering
	\caption{The number of parameters for deployment / for continued learning in thousands for our method and several existing methods.}\label{tab:cmpx_new}
	\begin{tabular}{l|l|l|l|l}
		\# d/t param.[thousands]& \textbf{HalfCheetah-v2} & \textbf{Ant-v2} & \textbf{Hopper-v2} & \textbf{Walker2D-v2}   \\
		\hline
		SAC 			& 74 / 289   & 100 / 389  & 71 / 280  & 74 / 289 \\
		SAC-Big 		& 1081 / 4300  & 1182 / 4695 & 1069 / 4264  & 1081 / 4301 \\
		OFE				& 174 / 821   & 187 / 799  & 139 / 651 & 143 / 663 \\
		OFE-Big			& 1286 / 6276 & 1057 / 5114  & 924 / 4573    & 934 / 4613   \\
		\hline
		\hline
		Prune-A			& 152 / 1466  & 242 / 1111 & 55 / 556  & 361 / 2187 \\
		Prune-B			& 28 / 390    & 73 / 576   & 15 / 246  & 67 / 586  \\
		Prune-C			& 21 / 236    & 41 / 266   & 8 / 102   & 50 / 389  \\
	\end{tabular}
\end{table}

Figure~\ref{fig:eval_score_curves} shows the cumulative reward of the evaluation runs (every 10k steps) during training. We use a moving mean filter with a window-size of $5$ to smooth the curves slightly for a better visual presentation.
Our method generally converges at the same rate as the baseline methods. In the case of the Hopper environment the validation curves show a slight instability at iteration $\expnumber{2}{5}$, which, however, does affect convergence. In case of the Cheetah environment we find that OFE and OFE-Big can exhibit repeated downward-spikes in the evaluation curves. Using our method and possibly due to the additional noise introduced by the $\Xi$ RVs before the $\Theta$ parameters have converged completely,  this behavior can be amplified and such spikes can appear periodically in the latter half of training; however they do not affect the final convergence and score.

\begin{figure}[t!]
	\centering
	\includegraphics[width=0.49\textwidth]{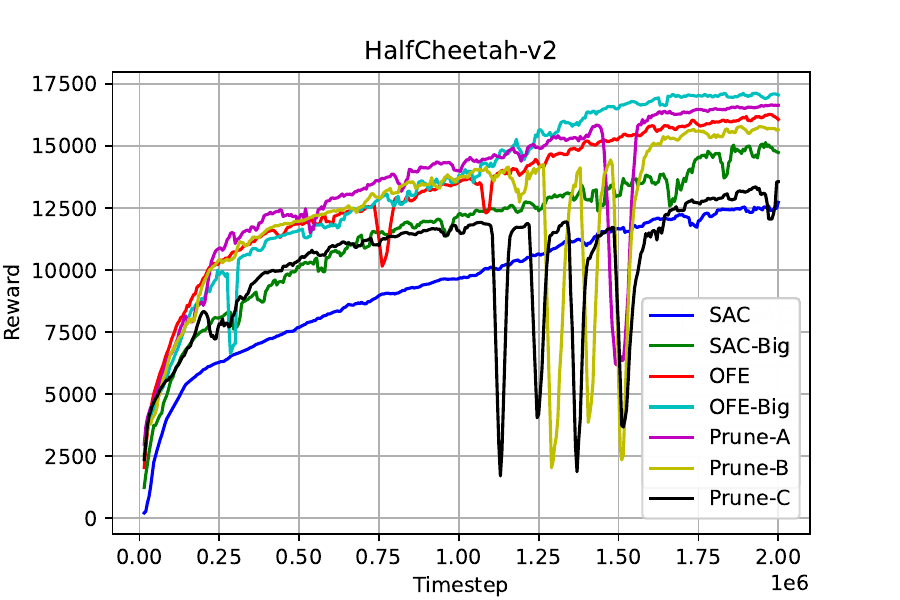}	
	\includegraphics[width=0.49\textwidth]{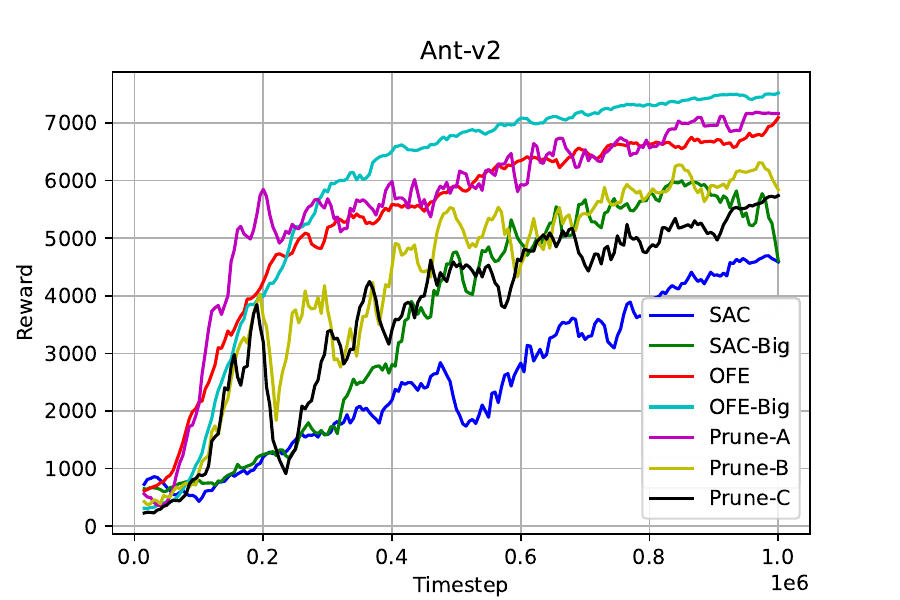}	
	\includegraphics[width=0.49\textwidth]{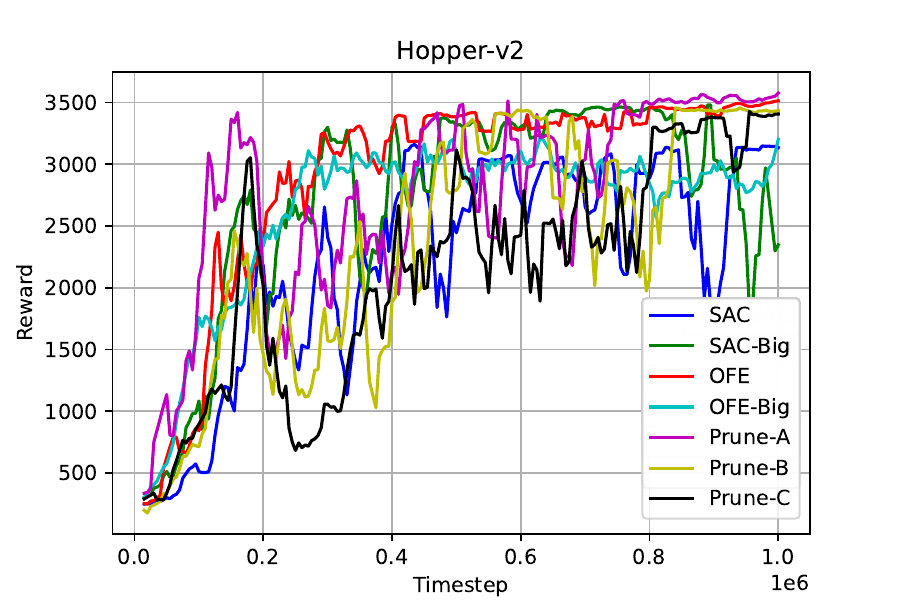}
	\includegraphics[width=0.49\textwidth]{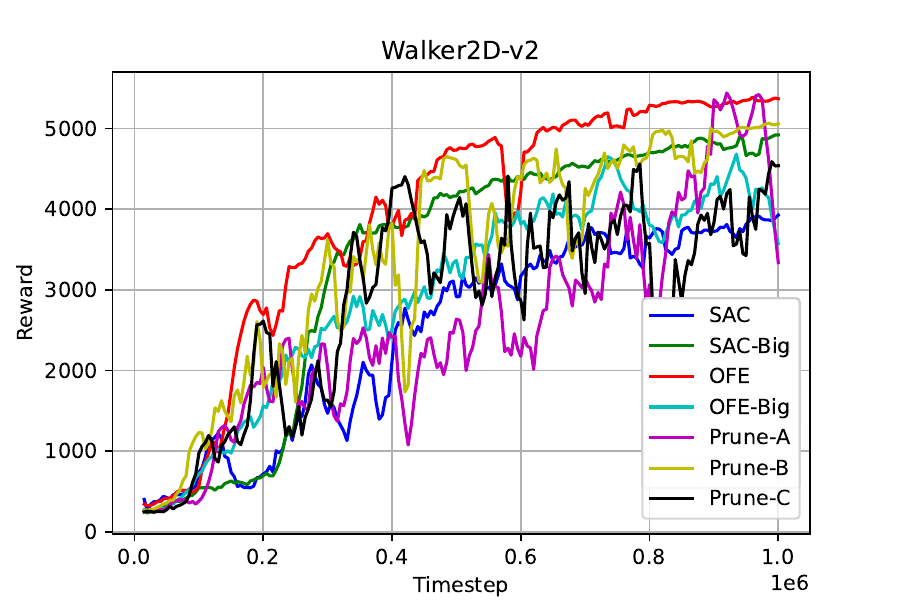}
	\caption{Cumulative rewards of the evaluation runs during training on the different environments. We compare our method using three sets of hyperparameters (Prune-A, Prune-B, Prune-C) to the standard SAC algorithm \cite{SAC_haarnoja2018soft}, SAC-Big using larger networks, the standard OFE algorithm \cite{ota2020can} and OFE-Big using larger networks.}\label{fig:eval_score_curves}
\end{figure}

Figure~\ref{fig:param_detail} shows in more detail the sizes of the found individual NN models with our method. The first number in each cell denotes the number of parameters in thousands in each model. Below this number is translated into a percentage with respect to the size of the individual networks of the OFE method on the left and with respect to the OFE-Big method's networks on the right. Again, a low number indicates that the found NN is small. Each tile is color-coded according to the first percentage. We observe that depending on the environment used, a different pattern is visible within each of the three plots. This means that our method can find different solutions suitable for each task. For example, in case of the Hopper task, the OFENet $\phi_{o,a}$ is consistently small (bright), while $\phi_{o,a}$ in the Cheetah and Walker tasks remains relatively much larger (darker).
The overall pattern when comparing the tiles of Prune-B with those of Prune-C remains similar. Prune-A uses a parameter $\rho=1.0$ instead of $0.5$ and hence, according to \eqref{eq:C-match}, does not focus on penalizing the size of the control policy and $\phi_{o}$ as much. As a result the policy $\pi$ remains relatively larger (darker).

\begin{figure}[t!]
	\centering
	\includegraphics[width=0.49\textwidth]{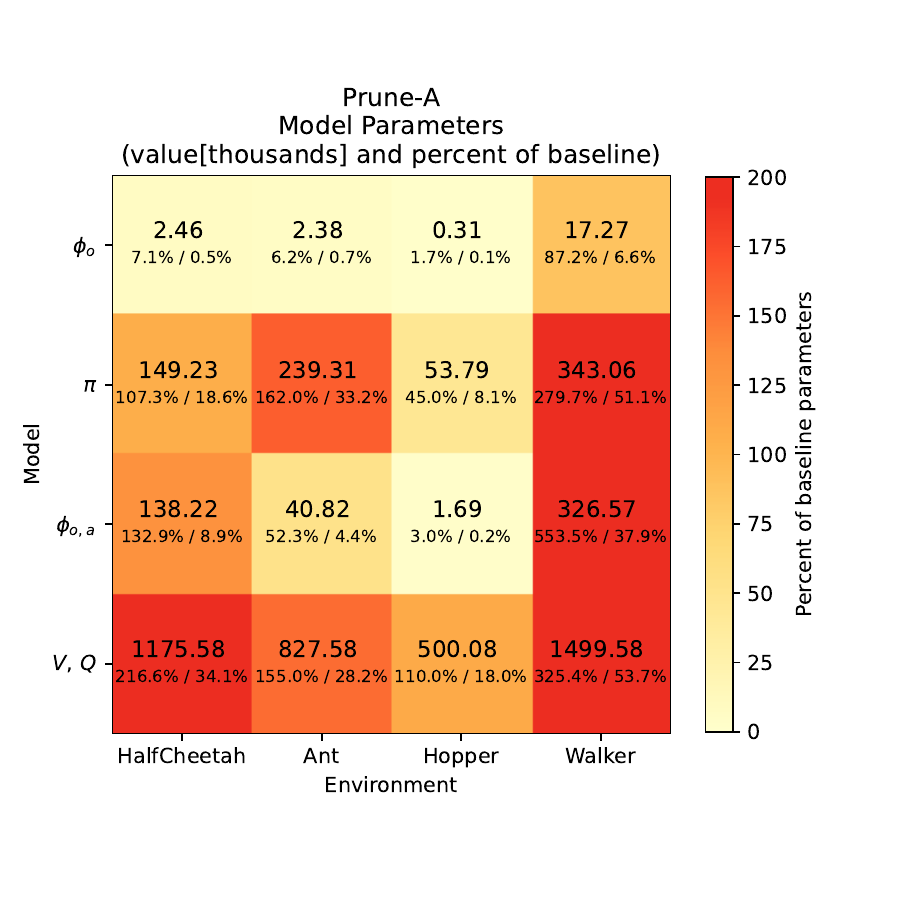}	
	\includegraphics[width=0.49\textwidth]{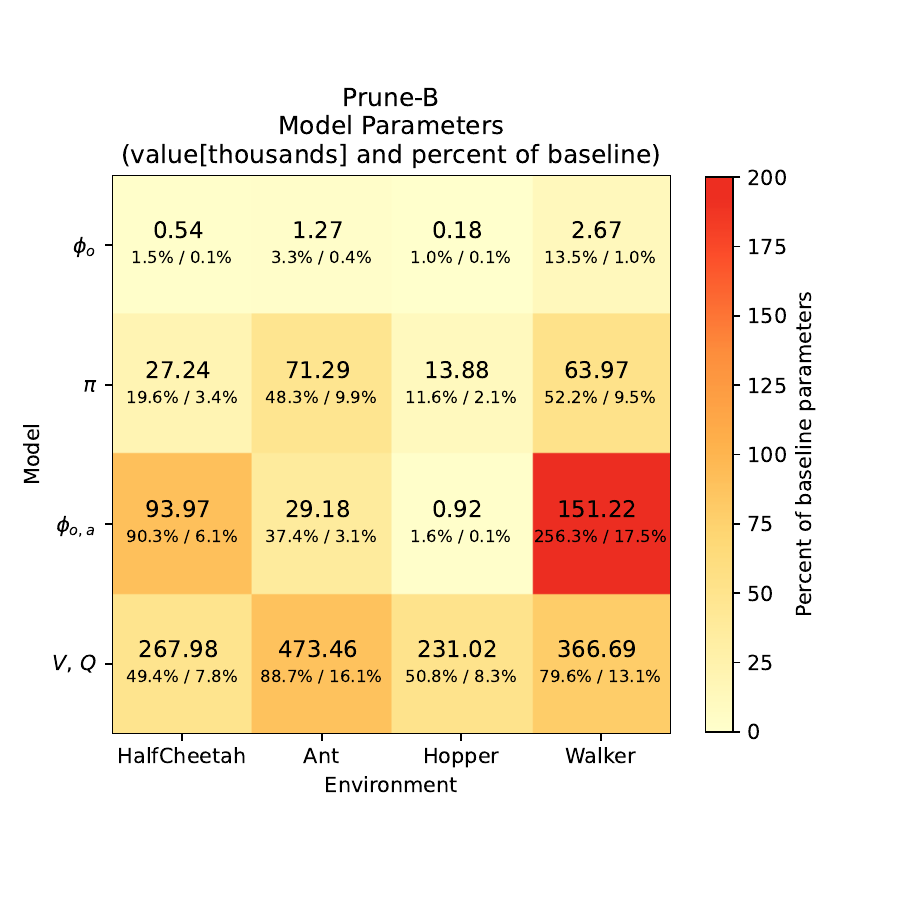}	
	\includegraphics[width=0.49\textwidth]{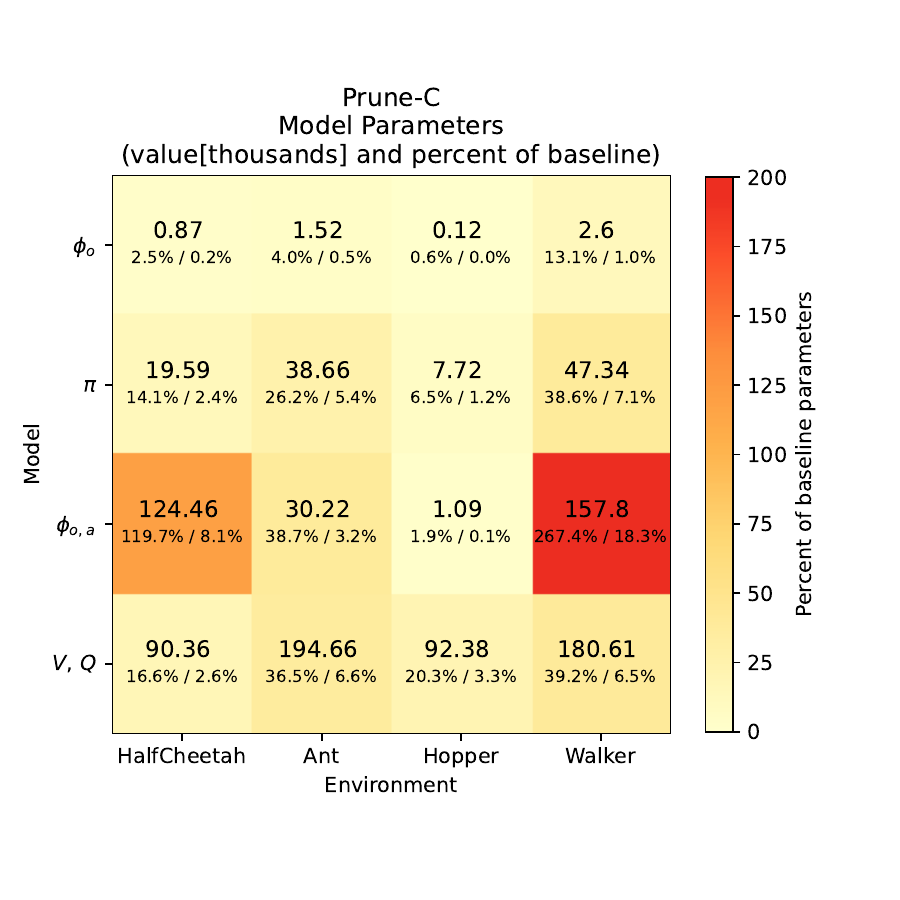}
	\caption{The number of parameters in each found network using our method in thousands. Below each number, it is translated to a percentage with respect to the OFE baseline's networks on the left, and with respect to the OFE-Big baseline's networks on the right. Each tile is color-coded according to the first percentage.}\label{fig:param_detail}
\end{figure}

Figure~\ref{fig:archs} shows the number of introduced units in each layer of the dense OFEXiNet after training and pruning for Prune-A, Prune-B, Prune-C and the baseline OFE-Big.
The first half of layers corresponds to $\phi_{o}$ and the second half to $\phi_{o,a}$. Over the bars in each half we sum up the units of those layers. These numbers are the dimensions by which the observation $o_t$ and the tuple $o_t,\, a_t$ are extended by the final OFEXiNet to yield $z_{o_t}$ and $z_{o_t,a_t}$, respectively.
OFE-Big uses for the Cheetah task an 8 layer $\phi_{o}$ OFENet and increases the dimension when going from $o_t$ to $z_{o_t}$ by $8\cdot128 = 1024$. Similarly an 8 layer $\phi_{o,a}$ is used and the dimension when going from $(z_{o_t},\, a_t)$ to $z_{o_t,a_t}$ is increased by  $1024$ as well.
Using our method, the found OFEXiNet architectures are task dependent. In all tasks the first part $\phi_{o}$ of the OFEXiNet is pruned heavily. The second part $\phi_{o,a}$ remains wider and most importantly, shows different patterns for each environment. In the case of the Cheetah (upper left) each layer introduces a similar number of units. Note that plot for the Hopper environment (lower left) is enlarged for better visibility and shows a slight decrease in the number of units towards the end of $\phi_{o,a}$.
In the case of the Ant environment (upper right) the later layers introduce more units, while we see the opposite behavior in the Walker2D (lower right) environment.

\begin{figure}[t!]
	\centering
	\includegraphics[width=0.49\textwidth]{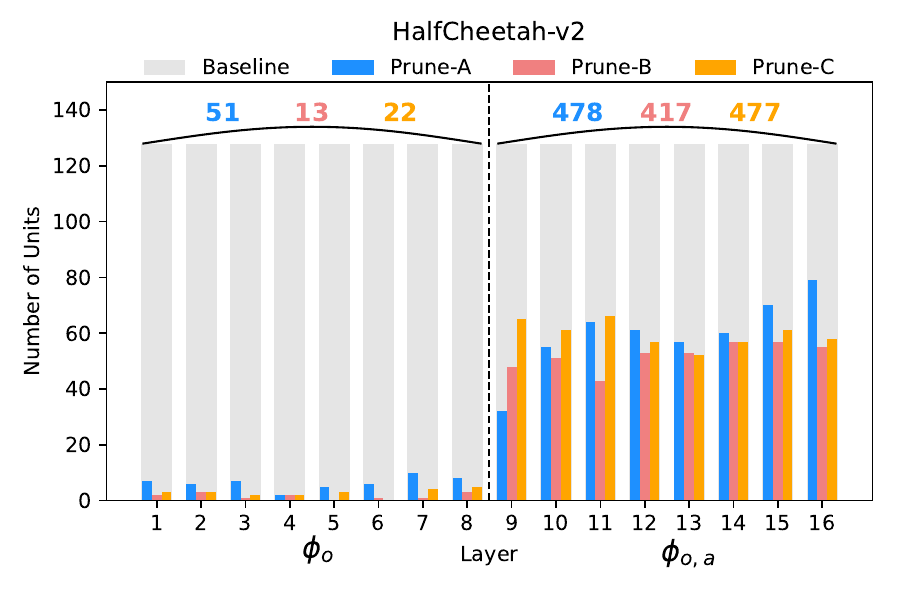}	
	\includegraphics[width=0.49\textwidth]{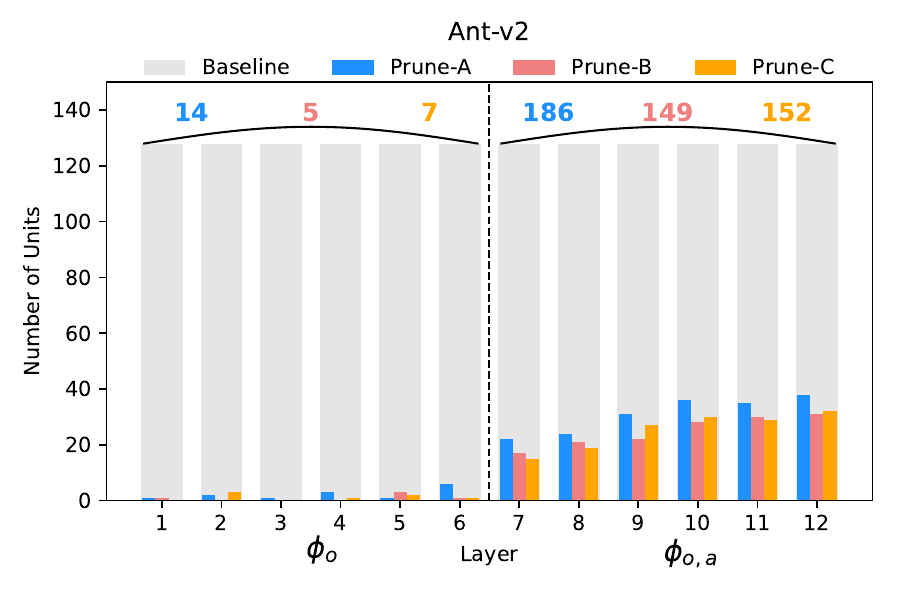}
	\includegraphics[width=0.49\textwidth]{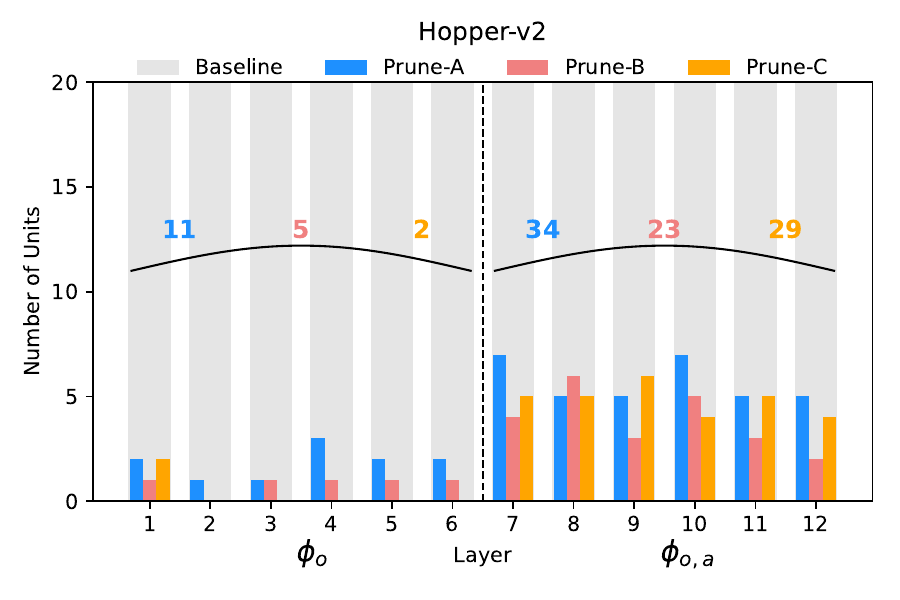}
	\includegraphics[width=0.49\textwidth]{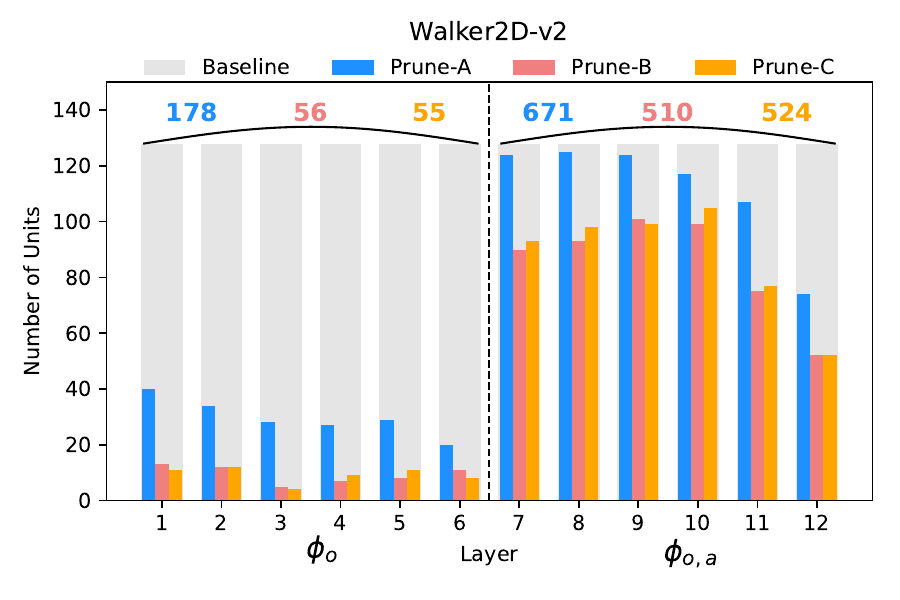}
	\caption{The number of units in each layer of the OFEXiNets after training and pruning for $\phi_{o}$ and $\phi_{o,a}$ on the left and on the right, respectively in each plot. Over the bars the sum of all units of $\phi_{o}$ and $\phi_{o,a}$ are stated. Note that the plot for the Hopper environment has been enlarged for better visibility.}\label{fig:archs}
\end{figure}

\section{Conclusion}\label{sec:conclusion}

We presented a systematic approach for integrating simultaneous training and pruning into modern reinforcement learning algorithms that employ large-scale deep neural networks, specifically those augmented with Online Feature Extractor Networks (OFENets) \cite{ota2020can}. Our method, OFEXiNet, relies on XiNets, first introduced in \cite{Guenter24_Robust}, which employ Bernoulli random variables that multiply the structures of a neural network considered for pruning and their posterior distribution of which is approximated via variational inference. The samples of such random variables realize different subnetworks during training, where any structure of the neural network can be active or inactive.

Parameter and computational efficiency are promoted by embedding carefully designed regularization terms from \cite{Guenter24_Robust} on the parameters of the random variables into the optimization cost function. This regularization guides the network to explore and retain only the most critical subnetworks during training, leading the Bernoulli random variables’ parameters to converge to either 0 or 1, resulting in a deterministic and pruned network. As a result, such networks dynamically reduce their complexity during training when a structure becomes inactive.

Inspired by \cite{guenter2024complexityawaretrainingdeepneural}, and central to our approach, is the definition of a joint objective that balances standard RL objectives with sparsity regularization terms tailored to the dense connectivity of the OFENet’s DenseNet architecture. We calculate the parameter complexity of the used OFEXiNets and XiNets of the RL algorithm in terms of the random variables’ parameters and show that, for a specific choice of prior parameters, the complexity cost becomes a part of several RL optimization problems underlying the learning of the neural networks.
By applying this framework throughout the entire RL training process, we enable agents to start learning with highly expressive, over-parameterized models and to adaptively prune redundant units in both the feature extractor and policy, as well as value networks of the agent.

We evaluated our method on popular continuous control environments (MuJoCo) and with the prominent RL agent SAC, confirming that, by using our approach, OFEXiNet, the large neural network models in RL algorithms using OFENets can be reduced down to $40\%$ of their initial size with only a slight performance loss. An even greater reduction is obtained for the networks required for the deployment of the agent, possibly on low-powered mobile devices. Our results show that pruning during training can also yield small, high-performing agents, outperforming traditional SAC agents at a lower total complexity of the involved neural networks.

\end{document}